\documentclass{article}

\PassOptionsToPackage{numbers, compress}{natbib}

\usepackage[preprint]{neurips_2020}

\usepackage{enumitem}

\usepackage{wrapfig}

\usepackage[utf8]{inputenc} 
\usepackage[T1]{fontenc}    
\usepackage{hyperref}       
\usepackage{url}            
\usepackage{booktabs}       
\usepackage{amsfonts}       
\usepackage{nicefrac}       
\usepackage{microtype}      

\usepackage{algorithm}
\usepackage{algorithmic}

\usepackage{graphicx}
\usepackage{amsmath,amssymb}
\usepackage{subfig}
\usepackage{wrapfig}

\graphicspath{{./}{./images/}{./images_supplementary/}{./arxiv/}}

\title{Constrained Combinatorial Optimization with Reinforcement Learning}

\author{%
  Ruben Solozabal\thanks{Corresponding author: ruben.solozabal@ehu.eus} \\
    Department of Communications Engineering\\
  University of the Basque Country\\
   Bilbao 48013, Spain \\
  \texttt{ruben.solozabal@ehu.eus} \\
   \And 
Josu Ceberio \\
   Department of Computer Science and\\ Artificial Intelligence\\
    University of the Basque Country \\
   Donostia 20018, Spain \\
   \texttt{josu.ceberio@ehu.eus} \\
   \AND
   Martin Tak\'a\v{c} \\
   Lehigh University \\
Industrial and Systems Engineering Department \\ 
200 West Packer Avenue\\
Bethlehem, PA 18015, USA \\
   \texttt{Takac.MT@gmail.com} \\
}

\begin{document}
\setlength{\abovedisplayskip}{3pt}
\setlength{\belowdisplayskip}{3pt}

\maketitle

\begin{abstract}

This paper presents a framework to tackle constrained combinatorial optimization problems using deep Reinforcement Learning (RL). To this end, we extend the Neural Combinatorial Optimization (NCO) theory in order to deal with constraints in its formulation. 

Notably, we propose defining constrained combinatorial problems as fully observable Constrained Markov Decision Processes (CMDP). In that context, the solution is iteratively constructed based on interactions with the environment. The model, in addition to the reward signal, relies on penalty signals generated from constraint dissatisfaction to infer a policy that acts as a heuristic algorithm.  Moreover, having access to the complete state representation during the optimization process allows us to rely on memory-less architectures, enhancing the results obtained in previous sequence-to-sequence approaches. Conducted experiments on the constrained Job Shop and Resource Allocation problems prove the superiority of the proposal for computing rapid solutions when compared to classical heuristic, metaheuristic, and Constraint Programming (CP) solvers.
\end{abstract}

\section{Introduction}
\label{introduction}

Combinatorial optimization is the science that studies finding the optimal solution from a finite set of discrete possibilities. Historically, combinatorial problems have been approached with exact algorithms, which guarantee the optimality of the solution. Conversely, they are not guaranteed to do it in polynomial time, and thus, when large problems are optimized, exact methods are no longer a feasible option. Today, many of the combinatorial problems are classified, according to the theory of complexity, as NP-Hard~\cite{gary1979computers}. For problems that fall in this category, it is intractable to achieve the optimal solution systematically, as an algorithm to solve them in polynomial time has not been discovered yet.
\\[3pt]
In those cases, approximation methods such as handcrafted heuristics are used to obtain near-optimal solutions. Nevertheless, designing such types of algorithms for combinatorial optimization can be a daunting task that requires expertise on the problem. Because of this, the recent idea to infer heuristics without human intervention is an appealing objective. As demonstrated in  \cite{bello2016neural}, Reinforcement Learning (RL) can be used to that achieve that goal. In the Neural Combinatorial Optimization (NCO) framework, a heuristic is parameterized using a neural network to obtain solutions for many different combinatorial optimization problems without hand-engineering. The only requirement is that evaluating the objective function must not be time-consuming.
\\[3pt]
In this paper, we extend the work presented by Bello \cite{bello2016neural}, introducing constrained combinatorial problems into the NCO framework. So far, constrained problems have not been approached using this technique. Only combinatorial problems in which the neural network can be set up to ensure feasible solutions have been addressed. To that end, neural models have limited the action space the neural network produces to avoid dealing with unfeasible results. Usually, implementing a masking scheme over the output probability distribution. In this manuscript, we deal with constraints introducing them as penalty coefficients into the reward objective function. This way, it is possible to deal not only with maskable constraints but also with constraints that cannot be evaluated during the resolution process, broaden NCO to general constrained combinatorial problems.
\\[3pt] 
To enhance this technology, we need better ways to infer a policy on the neural network. We also contribute to this sense. Unlike the original NCO proposal, where the solution is computed based on a single interaction with the environment, here, we consider the Constrained Markov Decision Process (CMDP) formulation for the problems. Therefore, the solution is generated as a sequence of decisions based on intermediate states obtained during the resolution process. This model uses the state representation to give the agent a better understanding of how the solution is evolving, thereby improving the quality of the results obtained.
\\[3pt]
To validate the proposed framework, we learn heuristics for two relevant and well-known operations research problems: Job Shop Scheduling (JSP)~\cite{garey1976complexity,chaudhry2016research} and Virtual Resource Allocation Problem (VRAP) \cite{beloglazov2012energy}. For both cases, constrained variants of the problems were considered. Conducted experiments point out that our model outperforms classical heuristic algorithms, metaheuristics, and the Constraint Programming (CP) solver CP-SAT from OR-Tools when real-time solutions need to be obtained. Moreover, the model shows a robust behavior, as the solutions' quality presents a low variance between different problem instances.

\section{Background}
\label{background}

The use of neural networks for solving combinatorial optimization problems dates back to \cite{hopfield1985neural}. The authors applied the Hopfield-network for solving instances of the Traveller Salesman Problem (TSP). Nevertheless, the application of neural networks on combinatorial problems was limited to small scale problem instances due to the available computational resources at that time. It has been in the last few years with the rise of deep learning that this topic has again attracted the attention of the artificial intelligence community.
\\[3pt]
Recently, \cite{vinyals2015pointer} trained a Deep Neural Network (DNN) to solve the Euclidean TSP using supervised learning. They proved that a neural network is able to parametrize a competitive policy also in domains with large action spaces as it is the case of most real-world combinatorial problems. To this end, they introduced the Pointer Network (PN), a neural architecture that enables permutations of the input sequence. Despite their positive results, using supervised learning to solve combinatorial problems is not trivial, as acquiring a training set implies the ability to solve a large number of instances optimally.   
\\[3pt]
In \cite{bello2016neural}, the NCO framework was presented, and RL was implemented for the first time to solve combinatorial problems. The authors took the Pointer Network introduced by Vinyals and utilized it in an actor-critic architecture to solve the TSP problem. The work proved that it is possible to learn competitive heuristics without human intervention. Although, they applied heavy sampling and searching techniques at interference to improve the solution generated by the neural network itself.
In \cite{deudon2018learning}, the TSP problem was also optimized using the NCO approach. On this occasion, a Transformer Network \cite{vaswani2017attention} is used, a top performance architecture in Natural Language Processing.
In that work, the greedy output of the neural network is hybridized with a local search to infer better results. Independently, \cite{kool2018attention} also presented the same architecture, yet with improvements in the decoder as well as in the training mechanism. Benefits that allows them to be competitive without applying search strategies at inference.
\\[3pt]
All the works presented so far have major similarities: they are based on sequence-to-sequence models, architectures originally designed for supervised learning. In these models, the solution is decoded at once based only on a single interaction with the problem. Conversely, \cite{nazari2018reinforcement} approached the Vehicle Routing Problem (VRP) as an MDP. Specifically, they built the solution as a result of a sequence of decisions made interacting with the environment. Such an approach allows to focus on how the environment evolves to construct the solution. This strategy also enables to deal with stochastic versions of the problem.
\\[3pt]
So far, the literature has used NCO to solve combinatorial problems without dealing with constraints. E.g.,~\cite{mirhoseini2017device, nazari2018reinforcement, mao2016resource, mao2019learning} address scheduling problems relying on masking schemes to avoid unfeasible actions during the resolution process. Nevertheless, this cannot be done for many combinatorial problems. A different approach could be~\cite{chen2019learning}. That work deals with combinatorial problems performing an RL-assisted local search procedure. However, this method a is only viable in case a feasible solution is easy to find. In this paper, we extend NCO to deal with constraints in its formulation, allowing under this framework to compute rapid solutions even in highly constrained problems.



\section{Neural Constrained Combinatorial Optimization}
\label{neural_constrained_combinatorial_optimization}

In this section, we formally define NCO for solving constrained combinatorial problems. Firstly, we introduce the terminology to use in this manuscript. Let us represent each instance of the problem as a static feature vector $s \in \mathit{S}$, where $S$ stands for the whole distributions of instances of the problem the model needs to learn a heuristic. This vector $s$ defines the instance and does not change during the interaction with the environment. Let $d$ be the vector that represents the state of the environment. $d$ depicts the dynamic part of the input and evolves iteratively as partial decisions are made. The concatenation of those feature vectors at a time-step $t$ represents the input to our model $\{x_t \doteq (s,d_t), t=0,1,..,T\}$.
\\[5pt]\noindent
{\bf Limitations of Action-masked Networks.}
\label{limitations_of_action_masked_networks}
Current models used in NCO have limitations dealing with constraints. E.g., Pointer Networks (PNs)~\cite{vinyals2015pointer} can solve problems that require to compute permutations over the inputs (e.g., the TSP and the Knapsack problem). However, they are not directly applicable to other combinatorial problems. Other approaches incorporate a problem-specific masking scheme to avoid the infeasibilities the problem produces. It is the case  of~\cite{nazari2018reinforcement} for solving the VRP, the cities previously visited are masked to avoid selecting them later in the decision process.
\\[3pt]
Using masking schemes forces the neural model to produce feasible solutions and, therefore, ensures that the environment can evaluate the solution (and provide a reward signal). Nevertheless, masking schemes cannot be applied to all constraint problems. 
{\it It can be used only in problems in which the construction produces a valid solution.} However, for general combinatorial problems with constraint equations, which feasibility can be verified only at the end of the episode (when the solution is obtained) is a hard task. 
Therefore, in this work, we address these constrained problems defining them as Constrained Markov Decision Processes (CMDP)~\cite{altman1999constrained}. In this framework, the environment provides a reward signal and penalty signals generated from constraint dissatisfaction. This way, constraints can be incorporated as penalty terms into the objective function~\cite{borkar2005actor} and guide the agent to achieve feasible solutions.
\\[3pt]
\textbf{Remark 1:} Dealing with constraints as penalties is a key point in highly constrained environments. Providing bad rewards to unfeasible solutions can flatten the objective function, which leads to a lack of information to infer a competitive policy. Without this relaxation technique, it would be near impossible for the agent to achieve a feasible region, as it would not experience enough positive rewards. The problem of sparse reward is well known in RL~\cite{jaderberg2016reinforcement}. 
\\[5pt]\noindent
{\bf Reward constrained policy optimization.}
\label{reward_constrained_policy_optimization}
Reward constrained policy optimization method \cite{tessler2018reward, paternain2019constrained} requires a parametrization of the policy, as it is over the objective expected reward function where the penalty is added. In particular, we resort to Policy Gradients to learn the parameters of the stochastic policy $\pi_\theta(y|x)$ that, given as input the tuple composed by the instance of the problem and the state of the environment $x_t=\{s,d_t\}$, assigns high probabilities to actions $y$ that produce solutions with high reward, and low probabilities to those that do not. In Policy Gradients, the objective function $J_R^\pi(\theta)$ is defined as the expected reward for the policy $\pi$
\begin{equation}
\label{eq:1}
  J_R^\pi(\theta) =   \textstyle{\mathbb{E}}_{\tau \sim \pi_{\theta}}[R(\tau)].
\end{equation}
Non-maskable constraints are incorporated into (\ref{eq:1}) using the Lagrange relaxation technique. This allows us to shape the objective function, proportionally penalizing those policies that lead to infeasibilities. But first, for each constraint signal $C_i$, we define its expectation of dissatisfaction associated to the policy $\pi$ as
\begin{equation}
  J_{C_i}^\pi(\theta) =   \textstyle{\mathbb{E}}_{\tau \sim \pi_{\theta}}[C_i(\tau)].
\end{equation}
The primal problem becomes then to find the policy $\pi$ that maximizes the expected reward subject to the satisfactions of the constraints
\begin{equation}
\label{eq:9}
   \textstyle{\max}_{\pi \sim \Pi} J_R^\pi(\theta) \quad  \textrm{ s.t. } \quad J_{C_i}^\pi \leq 0 \quad \forall i.
\end{equation}
Using the Lagrange relaxation technique \cite{bertsekas1997nonlinear}, the problem statement in (\ref{eq:9}) is reformulated as an unconstrained problem where the unfeasible solutions are penalized. The objective function is therefore defined as
\begin{equation}
\label{eq:11}
g(\lambda)=  \textstyle{\max}_{\theta} J_L^\pi(\lambda, \theta)  =   \textstyle{\max}_{\theta} [J_R^\pi(\theta) - \textstyle{\sum}_{i} \lambda_i \cdot J_{C_i}^\pi(\theta)]  = \max_{\theta} [J_R^\pi(\theta) - J_{\xi}^\pi(\theta)],
\end{equation}
where $J_L^\pi(\lambda, \theta)$ denotes the Lagrangian objective function, $g(\lambda)$ stands for the Lagrange dual function, and $\lambda_i$ are the Lagrange multipliers, i.e., penalty coefficients. In this equation, we introduce the term $J_{\xi}^\pi(\theta)$ that defines the expected penalization, computed as the weighted sum of all expectation of constraint dissatisfaction signals.

It is noteworthy that setting the Lagrange coefficients $\lambda_i$ is a multi-objective problem where there exists a different optimum solution for each configuration. In this sense, we perform a manual selection of the penalty coefficients, although the optimal value can also be obtained using other alternatives, e.g., a multi-time scale learning \cite{tessler2018reward}.



The gradient of the Lagrangian objective function $J_L^\pi(\theta)$ is derived using the log-likelihood method. This derivation process is similar as deriving the expected reward, method introduced in \cite{williams1992simple}. The resulting gradient equation is
\begin{equation}
    \nabla_{\theta}J_L^\pi(\theta) =   \textstyle{\mathbb{E}}_{\tau \sim \pi_{\theta}(\cdot|s)}[(L(y|x) \cdot \nabla_{\theta} \log\pi_{\theta}(y|x)], 
\end{equation} 
where
$  L(y|x)  = R(y|x) - \xi(y|x)  
             = R(y|x) - \textstyle{\sum}_{i} \lambda_i \cdot C_i(y|x)
$ 
denotes the penalized reward obtained in each iteration, 
calculated subtracting from the reward signal $R(y|x)$ the weighed sum of all the constrained dissatisfaction signals $C(y|x)$. Lastly, the gradient is approximated via Monte-Carlo sampling, where $B$ problem instances are drawn from the problem distribution $s_{1},s_{2},\ldots,s_{B} \sim \mathcal{S}$. To reduce the variance of the gradients, and therefore, to speed up the convergence, we include a baseline estimator $b(x)$. Finally, the gradient of the Lagrangian function is defined as
\begin{equation}
    \nabla_{\theta}J_L^\pi(\theta) \approx \tfrac{1}{B} \textstyle{\sum}_{j=1}^{B} (L(y_{j}|x_{j}) -b(x_{j})) \cdot\nabla_{\theta} \log\pi_{\theta}(y_j|x_{j}).
\end{equation}
\noindent
{\bf Self-competing baseline estimator.}
\label{baseline_estimator}
The baseline function $b(x)$ estimates the reward the model achieves for a problem input $x$, such that the current result obtained for the instance $L^\pi(y|x)$ can be compared to the performance of 
$\pi$. The baseline estimator can be as simple as a moving average $b(x) = M$ with decay $\beta$, where $M$ equals $L^\pi$ in the first iteration, and updates as $M \leftarrow \beta M + (1-\beta) L^\pi$ in the following ones. A popular alternative is the use of a learned value function or critic $\hat{v}(x, \theta_{\nu})$, where the parameters $\theta_{\nu}$ are learnt from observations \cite{grondman2012survey}. A baseline estimator performs in the following way, the advantage function $L^\pi(y|x) - b(x)$ is positive if the sampled solution is better that the baseline, causing these actions to be reinforced, and vice-versa. 
\\[3pt]
Here, we propose a new baseline based on estimations over the current stochastic policy. This method allows us to not rely on an additional estimator for computing the baseline. In this approach, we increase the learning batch $B$ introducing $N$ times every instance of the problem. Since during the learning process the policy is stochastic, for every instance we obtain $N$ different solutions. This creates the reward distribution $Q_j^N$ we use the to estimate the current performance of the model on the instance $x_j$. In particular, we select the baseline estimator $b(x)$ as the quantile (e.g.: $\alpha=0.1$) of the obtained distribution. The baseline is therefore calculated as
\begin{equation}
    b(x_j)=\{q_j: Pr(Q_j^N \leq q_j) = \alpha\}.
\end{equation}


\section{Neural Network Model}
\label{neural_network_model}

In this paper, we argue that sequence-to-sequence models~\cite{sutskever2014sequence} compute the solution for combinatorial problems without interacting with the environment. Those models receive an instance of the problem and build a solution based only on the hidden state the decoder stores. By contrast, the proposed neural network presents similarities with traditional RL models used for solving fully observable MDPs. In those cases, computing the action distribution is not required to store any previous information, and therefore, memory-less architectures can be used for this purpose.
\\[3pt]
\textbf{Remark 2:} We argue that combinatorial problems can be defined as a fully observable CMDP. Since the solution is iteratively built interacting with the problem, partial solutions can be evaluated to give the agent a reference on how the solution evolves on the problem. In that perspective, the Recurrent Neural Network (RNN) used in the decoder on sequence-to-sequence models can be substituted by a memory-less DNN. This benefits the results as accessing the fully observable state of the problem is more reliable than doing on memories.
\\[5pt]\noindent
{\bf The proposed model.}
\label{proposed_model}
The model consists of two main components: an embedding and encoding part and a DNN in charge of computing the output distribution. The encoder codifies the instance of the problem $s$, which may be formed by sequences of different lengths. This corresponds to the static part of the input to our model and does not change during its resolution of the problem. The vector $s$ is combined with the state of the environment $d_t$ to create the input $x_t=\{s,d_t\}$ from which a DNN computes the action distribution. This part of the model is required to be computed in every interaction. It constitutes, therefore, the dynamic part of the model, and it is evaluated until the whole solution is completed.
\\[5pt]\noindent
{\bf State-based attention mechanism.}
\label{attention_mechanism}
Having access to state representation, allows to build a \textit{glimpse} mechanism over the state representation $d_t$. In this sense, this state-based attention mechanism extracts key features from the state representation $\hat{d}_t$ and uses this information to create a context vector $c_t$~\cite{bahdanau2014neural, luong2015effective} that introduces key information deeper into the model.


\section{Experimentation}
\label{experimentation}

To validate the proposed framework, we optimize two relevant and well-known constrained combinatorial problems: a Job Shop Problem~\cite{garey1976complexity,chaudhry2016research} and a Resource Allocation Problem~\cite{beloglazov2012energy}. In both cases, there exist a huge number of variants in the literature. For the sake of evaluating the potential of the proposed model, we select variants of the problems above that present both types of constraints we argue in this paper: constraints that can be embedded into the model and constraints that need to be relaxed and incorporated into the objective function.

\subsection{Job Shop Problem with limited idle time}


In the Job Shop Problem (JSP) there exist a number of $n$ jobs  $J=\{J_0,J_1...J_{n-1}\}$ and a set $m$ machines $M = \{M_0,M_1...M_{m-1}\}$. Within every job $J_i$ there is a number of operations $O_i$ that need to be processed in a specific order $O_i = \{O_{i,0},O_{i,1}...O_{i,m-1}\}$. For each operation $O_{i,j}$, the machine $M_{i,j}$ and the duration time $D_{i,j}$ associated are defined. The aim of this problem consists of assigning the jobs to the machines such that the operation period is minimized, also known as the \textit{makespan}. The classical JSP presents two types of constraints:
\begin{itemize}[noitemsep, topsep=0pt, label={-}]
  \item Precedence constraints: specify that for every two consecutive operations in a job, the first one must be completed before the second one can be scheduled.
  \item No overlap constraints: these constraints arise from the fact that a machine can only work in one operation at a time.
\end{itemize}
These constraints can be managed via a masking scheme, so we implement them as hard-constraints in our model. To include non-maskable restrictions, the JSP variant with {\it limited idle time} was considered. Under that constraint, for any machine, the period between finishing operation and starting the next operation (idle time) cannot exceed a certain threshold $T_{th}$. This constraint arises naturally in a real context, as the aim is usually to maximize the productivity of the machinery.
\\[3pt]
In the JSP with limited idle time, the objective function can be penalized by the sum of all intervals in which the idle time between operations exceeds the threshold $T_{th}$. Hence, the objective function to minimize is defined as
\begin{equation}
  L = \max {\mathcal{M}_i} + \lambda \textstyle{\sum}_{i,j}((t^{start}_{O_{i,j}} - t^{end}_{O_{i,j-1}})-T_{th})^+ ,
\end{equation}
where $\mathcal{M}_i$ denotes the time until the job $J_i$ is finished, and $t^{start}_{O_{i,j}}$ and $t^{end}_{O_{i,j}}$ the start and ending time scheduled for the operations.

\paragraph{Particularized model.}
\label{particularized_model}

The particularized neural model for the JSP problem is depicted in Fig.~\ref{fig:2}. The model at each time-step $t$, computes a binary action deciding whether the next operation for each job is scheduled. To this end, the model stores an index vector $i_t$ pointing at the operations that are required to be scheduled next. Remember that the operations for a job must be assigned in a specific order; that is operation cannot be scheduled until the previous one has finished. This procedure is repeated until all operations are assigned.

\begin{wrapfigure}{r}{0.5\textwidth}
\vskip -25pt
$\ $
\begin{center}
\centerline{\includegraphics[width=0.5\columnwidth]{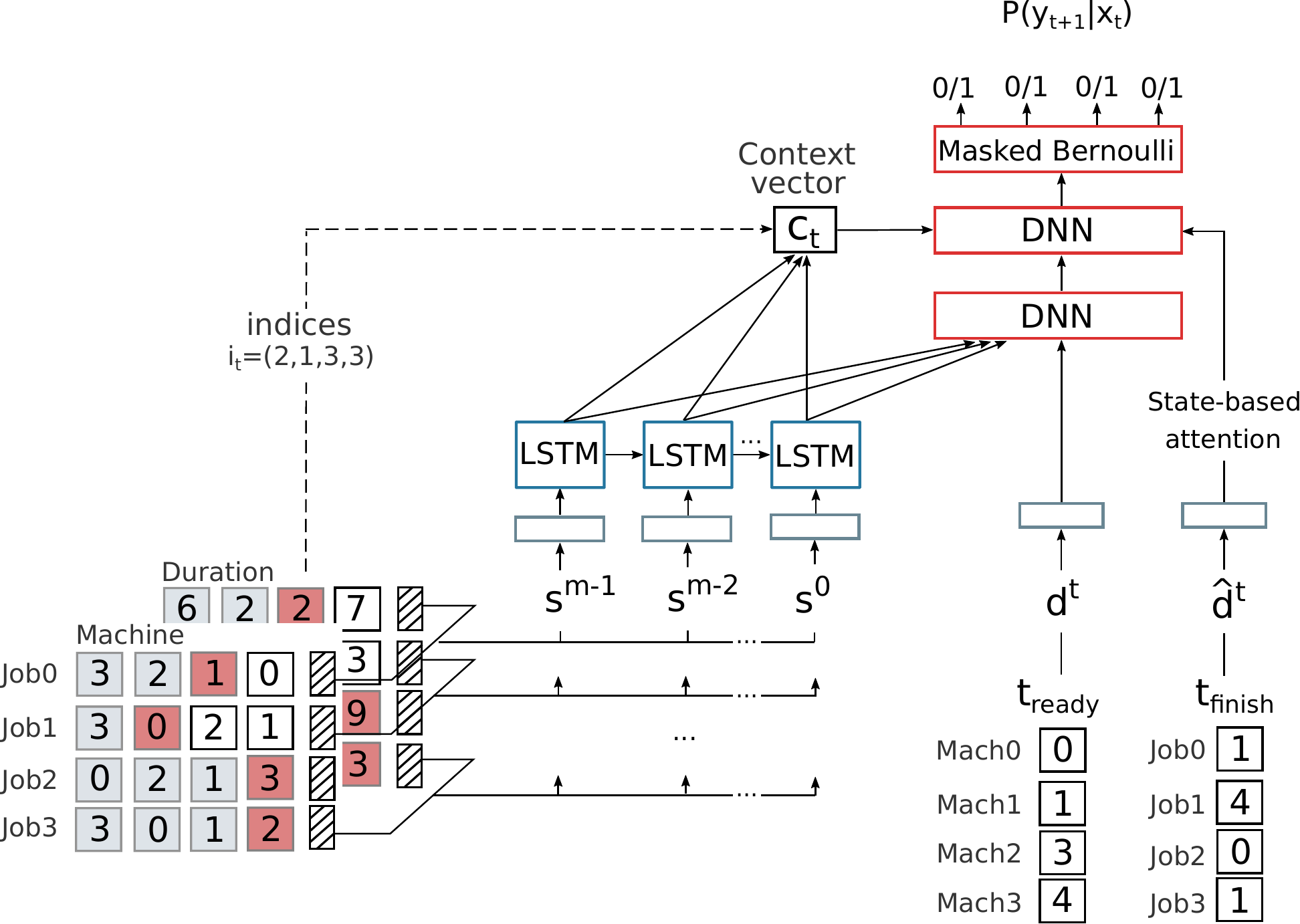}}
\caption{Neural model particularized for the JSP problem. The model is formed by a single LSTM encoder that operates over the sequence of operations for each job. The resulting vector that describes the combinatorial problem is combined with the state of the environment to decide the operations to be scheduled at each time-step $t$.}
\label{fig:2}
\end{center}
\vskip -0.38in
\end{wrapfigure}

As introduced, an instance of the JSP problem is defined 
by the machine assignation $M_{ij}$ and the time duration $D_{ij}$ matrices. For each operation $O_{ij}$, these values are concatenated to create the static input, denoted as $s_{ij}$ in the paper. This vector is embedded and sequentially encoded. In this case, the encoding process is configured backward for this problem. Producing, therefore, for each operation, a representation of the remaining operations until the job is completed. We refer to this vector as $e_{ij} = \mathit{enc}(s_{ij},..,s_{im}) \: \forall i$. 
\\[3pt]
The state of the problem is defined by the state of the machines and the operations currently being process at the decision time. We represent the state using two vectors: the first one indicates the number of time units until the machines are released, and the second one, the time left for the previous operation to finish. Those vectors constitute the dynamic part of the input $d_t$, and are recomputed at each time-step $t$.

\begin{wrapfigure}{r}{0.4\textwidth}
\vskip -20pt 
\centering
\footnotesize
\tiny 

JSP 10x10

     {\includegraphics[width=.4\textwidth,
    trim={0 10pt 35pt 20pt},clip]{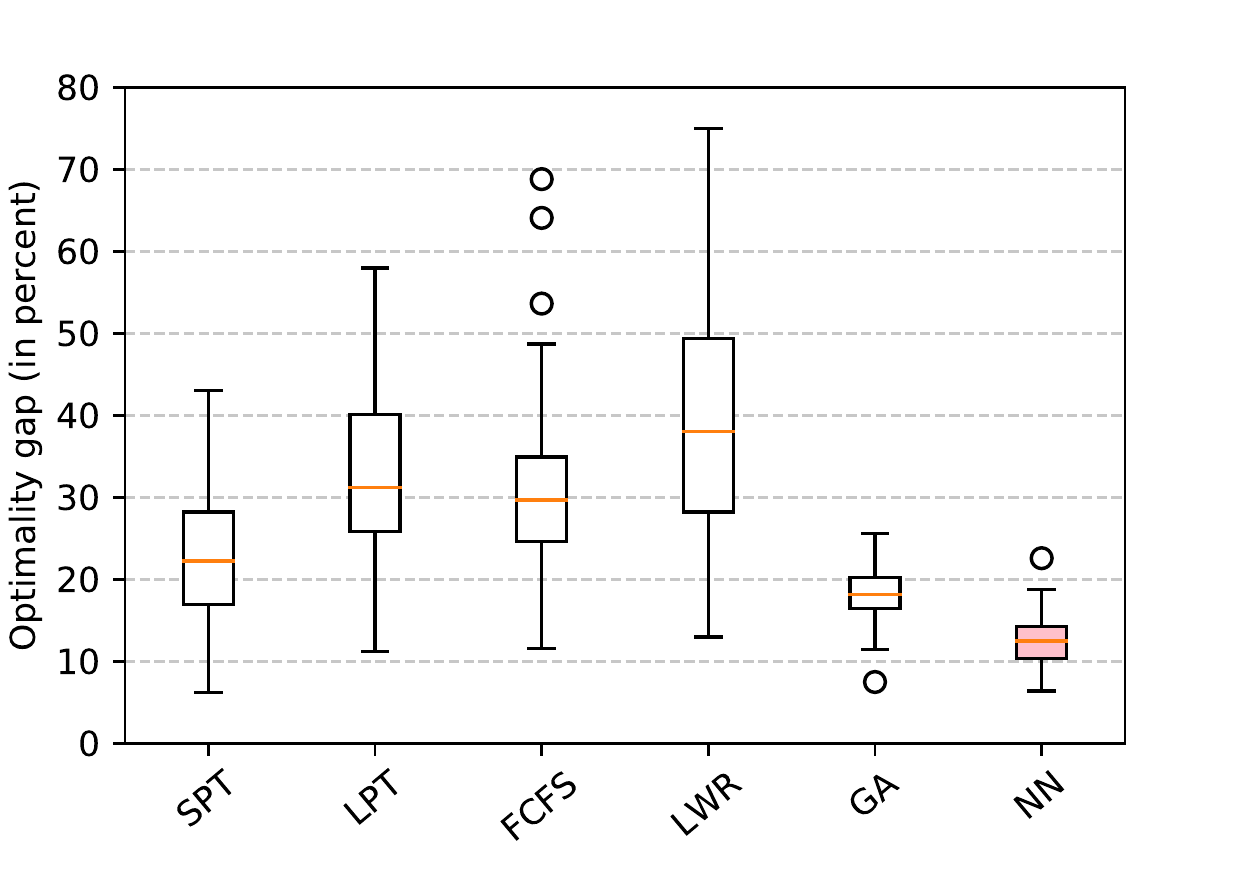}}
\vskip 3pt 
JSP 15x15
    
 {    \includegraphics[width=.4\textwidth,
    trim={0 10pt 35pt 20pt},clip]{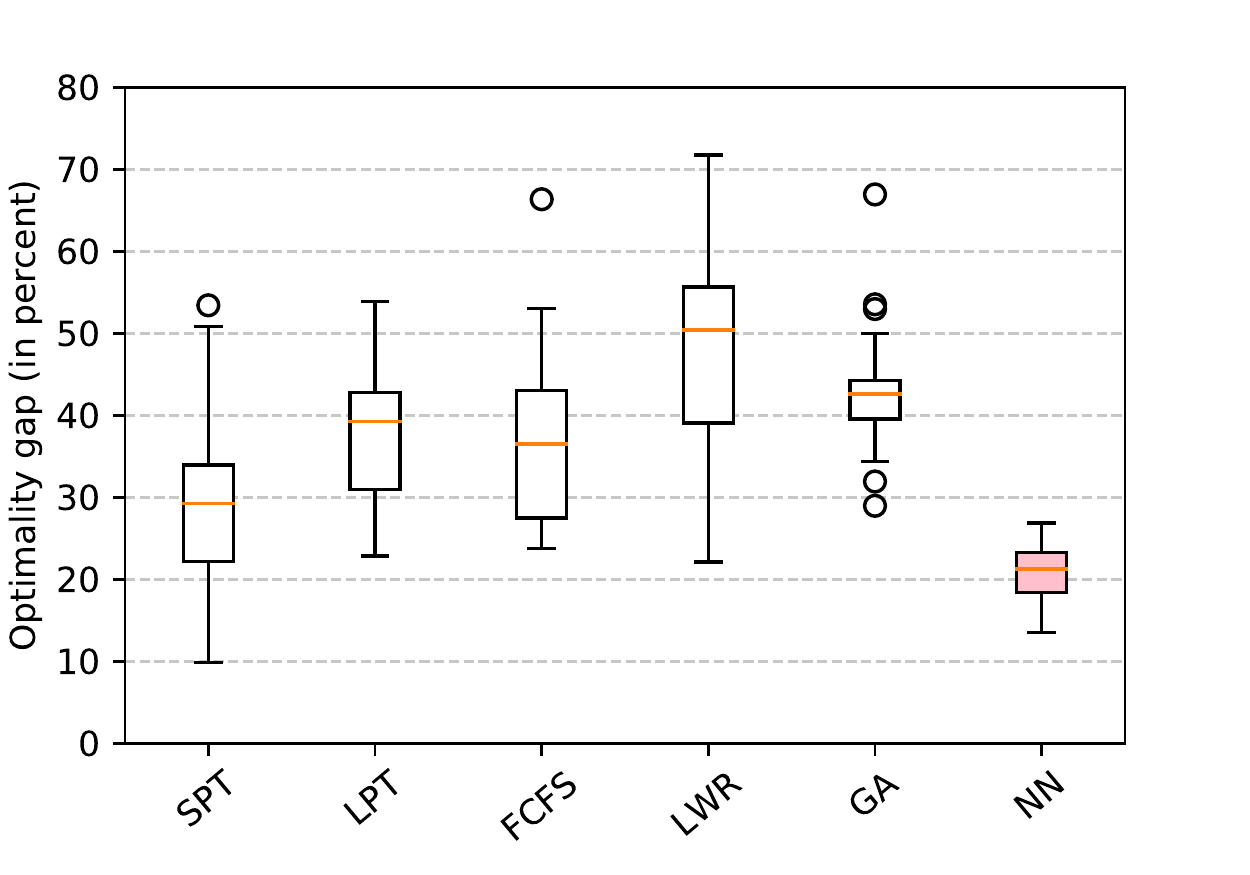}
    }
\vskip-10pt
  \caption{Comparison of the distance to the optimal solution in the \textit{classic} JSP between: different heuristics, a metaheuristic GA and our RL approach with sampling applied.}\label{fig:3}
  \vskip-40pt
\end{wrapfigure}

Both parts, the static and the dynamic state, are concatenated to create the input $x_t = (s,d_t)$ from where the DNN computes the output probability distribution (depicted in \textit{red} in Fig.~\ref{fig:2}). In this example, the output corresponds to a Bernoulli distribution, which indicates for each job whether the current operation (pointed by $i_t$) should be scheduled. Nevertheless, not every action can be selected at any time. Actions that lead to an infeasibility in the \textit{precedence} and \textit{no overlap constraint} are masked. This is achieved by forcing to zero the probability of scheduling the operation. In order to build the mask, the required information indicating whether the previous operation has finished or a machine is free to use can easily be gathered from the state vector $d_t$. 
\\[3pt]
Finally, the model presents a double glimpse mechanism: one on operations to be scheduled, and another over the state representation. The first one corresponds to the context vector $c_t$, generated gathering from $e_{ij}$ the indices pointed by the vector $i_t$. It acts as a glimpse on the operations yet to be scheduled in each job. The second one corresponds to the time for the previous operation to finish $\hat{d}_t$. Those vectors are introduced into the model, enhancing the features from which the solution is computed.

\subsection{Resource Allocation Problem}
\label{resource_allocation_problem}

In addition to the JSP, to prove the validity of the proposed framework, we evaluate its performance on the Virtual Resource Allocation Problem (VRAP)~\cite{beloglazov2012energy}. In this problem, a set of services is required to be allocated in a pool of server nodes. A service is composed of a chain of Virtual Machines (VM) that need to process a flow of information to fulfill a task. Here, the objective function to minimize is the energy cost of the entire set-up. This is calculated as the sum of the energy required to power up the servers plus the energy consumption of the virtual machines. Therefore, the problem is to arrange the services in the smallest number of nodes yet meeting the constraints associated with the infrastructure capacity and the service itself (e.g., maximum service latency). Further details on the problem can be found in Appendix B.

\begin{wrapfigure}{r}{0.4\textwidth}
\vskip -10pt
\centering

  \includegraphics[width=0.4\textwidth]{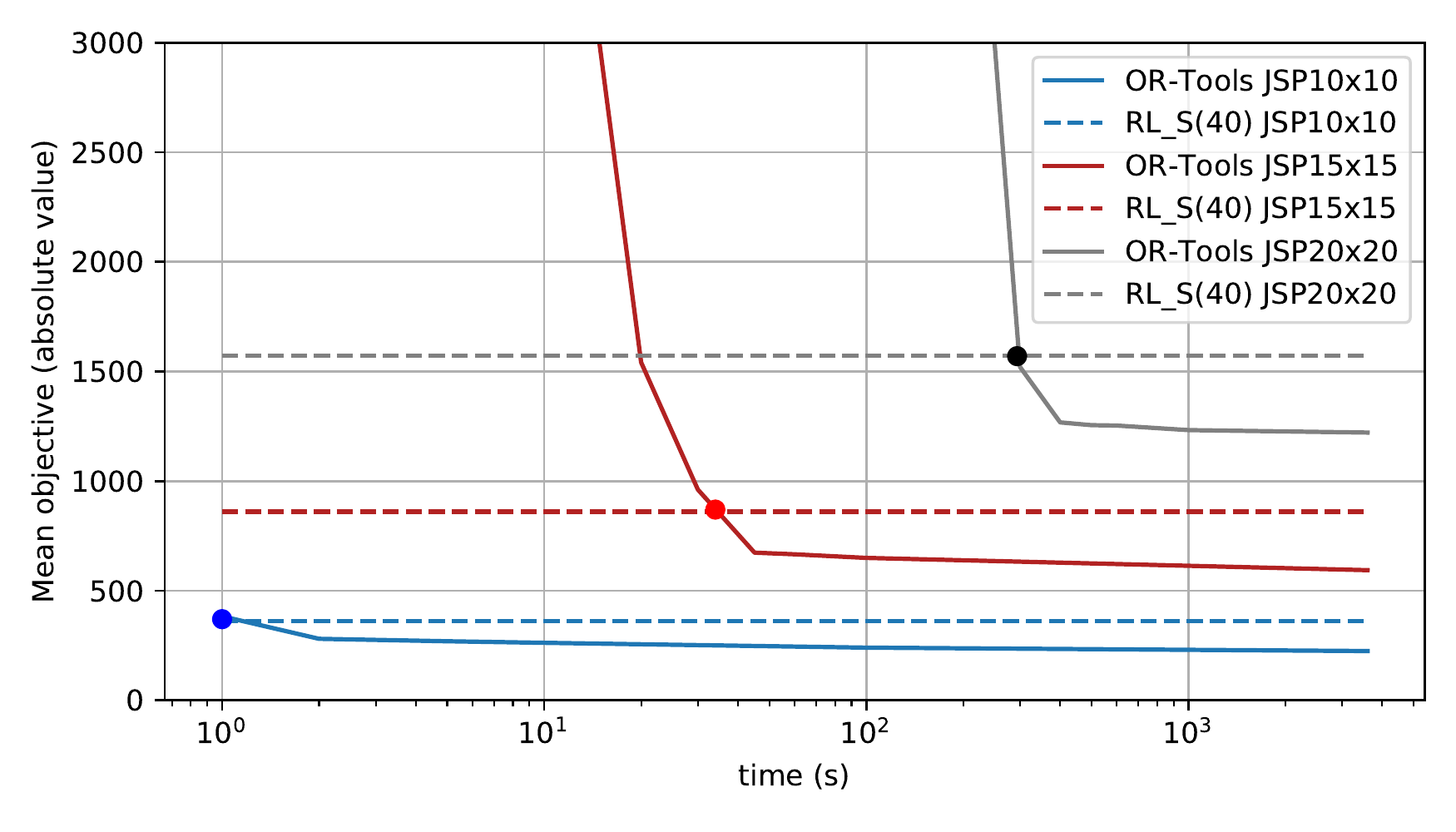}

  \caption{Mean objective value in function of the time obtained by OR-Tools for the JSP \textit{with limited idle time} ($\lambda=1$). The performance of the RL model is depicted in dashed line. The intersections between both representations are highlighted to indicate the time required for the solver to match the RL model.}
  \label{fig:4}
 
  \includegraphics[width=0.4\textwidth]{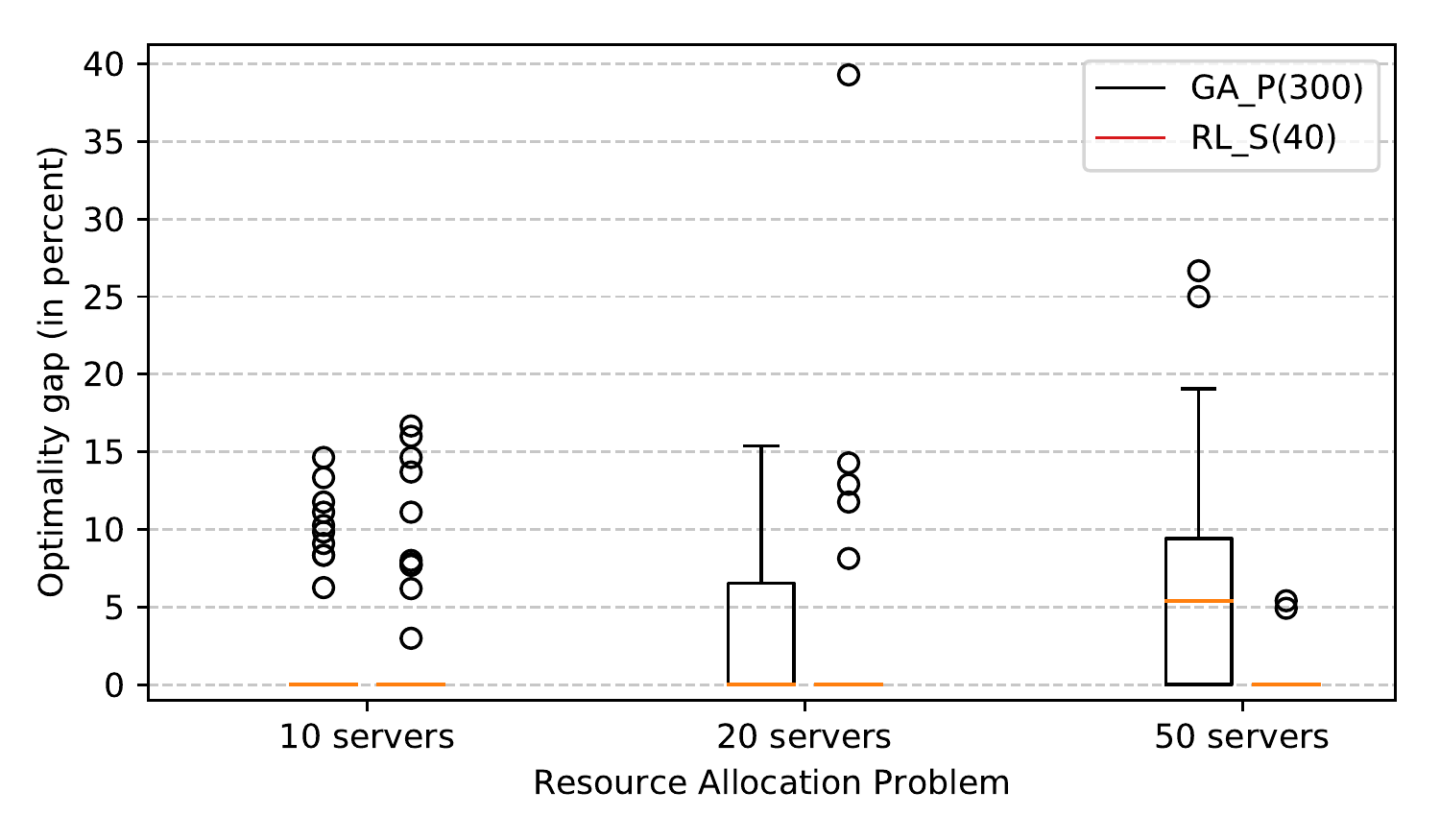}

  \caption{Comparison of the distance to the optimal solution in the Resource Allocation Problem between a Genetic Algorithm and our RL model.}
  \label{fig:5}
\vskip-30pt
\end{wrapfigure}

The RL model we propose to address this problem is similar to the previous one. The service composed by a sequence of VMs, each one represented by its specific features, is encoded using an RNN. The resulting vector represents the static part of our input. It is combined with the state of the environment to feed the neural network that iteratively decides the server in which each VM in the chain is going to be located. Here, the physical resource constraints can be guaranteed by the model. Nevertheless, the restrictions associated with the whole service (e.g., end-to-end latency) cannot be checked until the solution is commuted. Therefore, and as shown in the previous example, these constraints are relaxed and introduced as penalty terms into the objective function.

\begin{table*}[t]
\caption{Average objective, standard deviation and mean computing time for instances of the \textit{classic} JSP ($\lambda=0$) and JSP  \textit{with limited idle time} ($\lambda=1$). The size of the instance is denoted by the number of jobs $n$ and the number of machines $m$: JSP $n$x$m$.}
\label{tab:1}
\vskip -2pt
\centering
\scriptsize
\begin{sc}
\begin{tabular}{lcccccccccccc}
\toprule

Method      & \multicolumn{3}{c}{JSP10x10}                                      & \multicolumn{3}{c}{JSP15x15}                                       & \multicolumn{3}{c}{JSP20x20}                                       & \multicolumn{3}{c}{JSP25x25}                                       \\
\midrule
  ($\lambda = 0$)    & mean                 & std                  & time                 & mean                 & std                  & time                 & mean                 & std                  & time                 & mean                 & std                  & time                 \\
\midrule
SPT & \multicolumn{1}{r}{99.9} & \multicolumn{1}{r}{9.1} & \multicolumn{1}{r}{0.005s} & \multicolumn{1}{r}{153.2} & \multicolumn{1}{r}{10.5} & \multicolumn{1}{r}{0.012s} & \multicolumn{1}{r}{198.3} & \multicolumn{1}{r}{9.3} & \multicolumn{1}{r}{0.022s} & \multicolumn{1}{r}{252.9} & \multicolumn{1}{r}{13.2} & \multicolumn{1}{r}{0.045s}\\

LPT & \multicolumn{1}{r}{107.8} & \multicolumn{1}{r}{9.7} & \multicolumn{1}{r}{0.005s} & \multicolumn{1}{r}{163.9} & \multicolumn{1}{r}{10.9} & \multicolumn{1}{r}{0.012s} & \multicolumn{1}{r}{218.8} & \multicolumn{1}{r}{13.7} & \multicolumn{1}{r}{0.023s} & \multicolumn{1}{r}{278.2} & \multicolumn{1}{r}{17.2} & \multicolumn{1}{r}{0.050s}\\

FCFS & \multicolumn{1}{r}{107.1} & \multicolumn{1}{r}{10.0} & \multicolumn{1}{r}{0.005s} & \multicolumn{1}{r}{163.2} & \multicolumn{1}{r}{13.7} & \multicolumn{1}{r}{0.012s} & \multicolumn{1}{r}{219.3} & \multicolumn{1}{r}{13.1} & \multicolumn{1}{r}{0.023s} & \multicolumn{1}{r}{276.9} & \multicolumn{1}{r}{14.7} & \multicolumn{1}{r}{0.051s}\\

LWR & \multicolumn{1}{r}{113.9} & \multicolumn{1}{r}{14.1} & \multicolumn{1}{r}{0.037s} & \multicolumn{1}{r}{174.8} & \multicolumn{1}{r}{12.6} & \multicolumn{1}{r}{0.123s} & \multicolumn{1}{r}{227.3} & \multicolumn{1}{r}{12.7} & \multicolumn{1}{r}{0.279s} & \multicolumn{1}{r}{287.2} & \multicolumn{1}{r}{18.0} & \multicolumn{1}{r}{0.598s}\\

GA\_P(300)  & \multicolumn{1}{r}{96.4} & \multicolumn{1}{r}{5.2} & \multicolumn{1}{r}{55.80s} & \multicolumn{1}{r}{169.4} & \multicolumn{1}{r}{6.7} & \multicolumn{1}{r}{165.8s} & \multicolumn{1}{r}{254.3} & \multicolumn{1}{r}{7.2} & \multicolumn{1}{r}{303.9s} & \multicolumn{1}{r}{338.2} & \multicolumn{1}{r}{7.6} & \multicolumn{1}{r}{586.0s}\\

RL\_S(1) & \multicolumn{1}{r}{101.3} & \multicolumn{1}{r}{8.5} & \multicolumn{1}{r}{0.83s} & \multicolumn{1}{r}{161.9} & \multicolumn{1}{r}{11.9} & \multicolumn{1}{r}{2.15s} & \multicolumn{1}{r}{216.2} & \multicolumn{1}{r}{14.4} & \multicolumn{1}{r}{3.56s} & \multicolumn{1}{r}{277.2} & \multicolumn{1}{r}{17.3} & \multicolumn{1}{r}{5.16s}\\

RL\_S(40) & \multicolumn{1}{r}{91.9} & \multicolumn{1}{r}{5.8} & \multicolumn{1}{r}{1.04s} & \multicolumn{1}{r}{143.6} & \multicolumn{1}{r}{6.4} & \multicolumn{1}{r}{2.31s} & \multicolumn{1}{r}{196.9} & \multicolumn{1}{r}{7.7} & \multicolumn{1}{r}{4.38s} & \multicolumn{1}{r}{249.3} & \multicolumn{1}{r}{7.7} & \multicolumn{1}{r}{6.38s}\\

RL\_S(100) & \multicolumn{1}{r}{90.7} & \multicolumn{1}{r}{5.4} & \multicolumn{1}{r}{1.17s} & \multicolumn{1}{r}{142.1} & \multicolumn{1}{r}{6.6} & \multicolumn{1}{r}{2.65s} & \multicolumn{1}{r}{193.6} & \multicolumn{1}{r}{8.0} & \multicolumn{1}{r}{4.52s} & \multicolumn{1}{r}{244.5} & \multicolumn{1}{r}{8.1} & \multicolumn{1}{r}{7.04s}\\

OR-Tools & \multicolumn{1}{r}{81.5} & \multicolumn{1}{r}{4.6} & \multicolumn{1}{r}{0.082s} & \multicolumn{1}{r}{118.8} & \multicolumn{1}{r}{4.4} & \multicolumn{1}{r}{61.22s} & \multicolumn{1}{r}{156.2} & \multicolumn{1}{r}{4.5} & \multicolumn{1}{r}{1h(*)} & \multicolumn{1}{r}{195.4} & \multicolumn{1}{r}{4.9} & \multicolumn{1}{r}{1h(*)}\\

\midrule
 ($\lambda = 1$)     & mean                 & std                  & time                 & mean                 & std                  & time                 & mean                 & std                  & time                 & mean                 & std                  & time                 \\
\midrule

GA\_P(300)  & \multicolumn{1}{r}{343.7} & \multicolumn{1}{r}{45.5} & \multicolumn{1}{r}{91.6s} & \multicolumn{1}{r}{1117.0} & \multicolumn{1}{r}{39.0} & \multicolumn{1}{r}{257.3s} & \multicolumn{1}{r}{2476.0} & \multicolumn{1}{r}{62.6} & \multicolumn{1}{r}{578.4s} & \multicolumn{1}{r}{4453.2} & \multicolumn{1}{r}{111.7} & \multicolumn{1}{r}{1079s}\\

RL\_S(40) & \multicolumn{1}{r}{360.4} & \multicolumn{1}{r}{38.7} & \multicolumn{1}{r}{1.36s} & \multicolumn{1}{r}{860.2} & \multicolumn{1}{r}{65.4} & \multicolumn{1}{r}{3.12s} & \multicolumn{1}{r}{1573.0} & \multicolumn{1}{r}{112.7} & \multicolumn{1}{r}{5.18s} & \multicolumn{1}{r}{2745.0} & \multicolumn{1}{r}{173.1} & \multicolumn{1}{r}{7.75s}\\

OR-Tools & \multicolumn{1}{r}{221.4} & \multicolumn{1}{r}{18.4} & \multicolumn{1}{r}{1h(*)} & \multicolumn{1}{r}{593.5} & \multicolumn{1}{r}{19.9} & \multicolumn{1}{r}{1h(*)} & \multicolumn{1}{r}{1221.0} & \multicolumn{1}{r}{50.9} & \multicolumn{1}{r}{1h(*)} & \multicolumn{1}{r}{2075.0} & \multicolumn{1}{r}{101.2} & \multicolumn{1}{r}{1h(*)}\\

\bottomrule
\end{tabular}
\end{sc}
\begin{small}
\quad (*) The result is not optimal, the execution has been forced to end after the indicated time.
\end{small}
\vskip -0.1in
\end{table*}

\subsection{Results and analysis}
\label{results}

\paragraph{Job Shop Problem.}
\label{experimentation_job_shop}
We present the experimental study on the classical Job Shop Problem ($\lambda=0$) and also on the \textit{limited idle time} variant ($\lambda > 0$). The results obtained by our framework are compared with a Genetic Algorithm (GA)~\cite{cheng1996tutorial} and the solver CP-SAT from OR-Tools~\cite{CP}. In addition, in the case of the classic JSP results are also compared with some well-know heuristics: the \textit{Shortest Processing Time} (SPT),  \textit{Longest Processing Time} (LPT), \textit{First-come-first-served} (FCFS) and \textit{Least Work Remaining} (LWR)~\cite{mahadevan2010operations}. 
\\[3pt]
In the experimentation, two different decoding mechanisms are used in the RL model: a greedy and a sampling technique. In the greedy approach, the solution is directly obtained from the model, whereas in the sampling method, multiple solutions are computed from the stochastic policy, and the best one is selected. This comes naturally in this proposal with the self-competing strategy, therefore it does not add overhead to the model. These instances are referenced as RL\_S followed by the number of solutions taken in the experiment.
\\[3pt]
As noted, the results of the model are also compared with those obtained by OR-Tools. For small size instances, the solver is able to compute the optimal solution. However, for larger instances or when the number of restrictions is higher, as it is the case of the \textit{limited idle time} variant, computing the optimal solution becomes intractable. In those cases, we limited the execution time up to one hour, and the solutions obtained are only considered as near-optimal approximations. 
\\[3pt]
To implement the model, we use LSTM~\cite{gers1999learning} neural networks in the recurrent encoder and the objective function is optimized using Adam \cite{kingma2014adam}. The implementation details can be found in Appendix~A.
\\[3pt]
The results are summarized in \tablename~\ref{tab:1}. It introduces the average objective, the standard deviation and the mean computation time obtained by the different methods for the classic JSP $(\lambda=0)$ and the JSP \textit{with limited idle time} $(\lambda=1)$.  Performance measures were averaged on a set of 50 instances for each problem size. We observe that in the classic JSP, our approach is competitive in terms of the quality of the solution against the compared heuristics and the GA, especially for small and medium problems instances JSP10x10 and JSP15x15. Moreover, the variance obtained by the RL model is considerably low during the tests. We conclude, therefore, that the model is robust in the sense that the results are consistent in performance. It can also be observed that the sampling technique RL\_S(40) provides a reasonable tradeoff between the computational cost and the improvement in the results. It is, therefore, the number of samples we used hereof.
\\[3pt]
Fig.~\ref{fig:3} shows the optimality gap for the instances in which the optimal solution can be obtained in a reasonable time: JSP10x10 and JSP15x15. The optimality gap is defined as the difference (in percent) between the solution obtained and the optimum. According to the results, the RL\_S(40) outperforms the rest of the heuristics and metaheuristics. However, it presents a difference in performance when compared to the solver that goes from 11.2\% in the JSP10x10 up to 27.5\% in the JSP25x25 (see \tablename~\ref{tab:1}). 
\\[3pt]
Although the solver performed better than the RL model, the time required in each case is totally different, so carrying out a fair comparison is tricky. For this reason, we compute the time required by the solver to obtain a solution comparable with that of the RL model. We observe that in the case of the classical JSP ($\lambda=0$), the solver outperforms the other alternatives, and it does in a competitive computational time. Nevertheless, things turn around when the \textit{limited idle time} variant is considered. Although this problem adds no much complexity, the computation time required by OR-Tools increases significantly, in this case, a slight increase in the number of constraints in the problem is enough to prevent the solver from getting good approximations in the short time.
To illustrate that, we depict in Fig.~\ref{fig:4} the performance of the solver in function of the time elapsed. For the problem JSP15x15 and larger,  
the time required by the solver to match the performance of the RL model is orders of magnitude higher than the inference time. This becomes the RL model competitive for achieving rapid solutions.
\\[3pt]
{\bf Resource Allocation Problem.}
\label{resource_allocation_problem_experimentation}
The VRAP presents some differences when compared to the JSP. In this problem, services are located one at a time, and they are formed by sequences of no more than a few virtual machines. Also, the output is a categorical distribution over the servers in the infrastructure instead of binary decisions. These factors make it easier for the model to extract features from the problem definition. This point is reflected in a lower number of parameters of the neural network and faster training times. Further details on the implementation of the model can be seen in Appendix B.
\\[3pt]
In this problem, we compare the performance of the RL model with a GA and the OR-Tools CP solver. We depict the results in Fig.~\ref{fig:5}. In this case, we are able to compute the optimum with the solver, so the results are given relative to it. As shown in the picture, the RL model consistently predicts close to the optimal allocation sequences, outperforming the GA. In this case, the model is able to extract the features of the infrastructure and the services in order to infer a policy that almost suits perfectly on the problem instances.

\section{Discussion and Conclusion}
\label{discussion_and_conclusion}

In this work, we extend Neural Combinatorial Optimization (NCO) to include constrained combinatorial problems. We do it defining them as fully observable Constrained Markov Decision Processes (CMDP), where the proposed model iteratively constructs a solution based on the intermediate states obtained during the resolution process. To that end, in addition to the reward signal, the model relies on penalty signals generated from constraint dissatisfaction to guide the agent to achieve feasible solutions. This approach benefits from not requiring memory-based architectures to compute the solutions, which improves the quality of the results obtained.
\\[3pt]
The conducted experimental study points out that the proposed architecture presents the required versatility for being applied to real-world constraint combinatorial problems. Moreover, obtained results indicate that the RL model outperforms classical heuristics, metaheuristic, and CP solvers when real-time solutions need to be obtained.
\\[3pt]
Scaling the model for larger instances is an important direction for future research. We observe that the larger the combinatorial problem is, the harder it is for the model to infer good policies for the instances. The RL model tends to generalize well, but the performance gap becomes larger as the size of the problem increases.

\begin{ack}
This work was partially supported by the U.S. National Science Foundation, under award numbers
NSF:CCF:1618717, NSF:CMMI:1663256 and NSF:CCF:1740796, by the Basque Government Research Group IT1244-19 and by the Spanish Ministry of Science under the project TIN2016-78365-R.
\end{ack}

\bibliographystyle{abbrvnat}
\bibliography{ref}

\appendix

\section{Job Shop Problem}
\label{job_shop_problem}
In this appendix~\ref{job_shop_problem}, supplementary information on the Job Shop Scheduling Problem (JSP) is presented. Particularly, (1) details on the heuristic and metaheuristic algorithms included in the experimental study, (2) specifications on the implementation of the RL architectures for the JSP, and (3) running times of the learning process are introduced. 

\subsection{Heuristic and metaheuristic algorithms for the Job Shop Problem}
\label{heuristics}
Conducted experimental study in Section 5, compare the performance of the proposed model with some of the most representative heuristics and metaheuristic algorithms for the JSP in the literature. In the following, we present a summary of the four heuristics algorithms included, yet more information about them can be found in \cite{mahadevan2010operations}.

\textbf{Shortest Processing Time (SPT):} it is one of the most used heuristics for solving the JSP problem. At each iteration, it selects the job with the least processing time from the competing list and schedules it ahead of the others. With illustrative purposes, let us considerate that two operations of different jobs are competing at a time-step for the same machine to be released. In that case, the operation with the shortest processing time will be scheduled first.

\textbf{Longest Processing Time (LPT):} it follows the opposite rule of the SPT heuristic. The operation with the longest processing time is scheduled ahead of the competitors.

\textbf{First-Come-First-Served (FCFS):} this rule schedules the jobs simply in the order of job arrival. There is no consideration on the processing time or any other information. 

\textbf{Least Work Remaining (LWR):} it is also an extension of SPT, this rule dictates the operation to be scheduled according to the processing time remaining before the job is completed. The less work remaining in a job, the earlier it is scheduled.

In addition to the classical heuristic algorithms exposed above, a metaheuristic, particularly, a Genetic Algorithm (GA)~\cite{back1997handbook} has been included in the experimental study. The implementation corresponds to \cite{wurmen}. Regarding the hyperparameter setting, a population of 300 individuals, a crossover rate of 0.8 and a mutation rate of 0.3 were set. Finally, the algorithm was run 500 generations before stopping (enough iterations to converge in the different problems included in the study).

\subsection{Implementation details}

This appendix complements the details on the neural model introduced in Section 5.1. The proposed model presents two different input sources: the instance of the problem $s$, which is defined by the $M$ and $D$ feature matrices, and the state of the environment $d_t$, represented by the state of the machines and the time for the previous operations to finish. In order to embed both sources, single linear layers with a vector size of 64 are utilized. We find that normalizing the input vectors and embedding them in a higher feature space yields to superior solutions.

In regard with the details on the architecture, the RNN encoder used to codify the sequences of operations for each job is a single LSTM \cite{gers1999learning} layer with a hidden state size of 64. Specifically, it is a unidirectional encoder working backwards. This manner, the encoder outputs the codification of the remaining operations for a job, starting at every point in the sequence. This procedure is computed once and stored to be used during the interaction with the environment. In that process, an index vector $i_t$ points at the current operation to be scheduled and the feature vector $e_{ij}$ is gathered for each job to create the context vector $c_t$.

Lastly, the DNN decoder consists in multiple dense layers with a ReLU activation. The variables are initialized with Xavier initialization~\cite{glorot2010understanding}. The batch size is 800, and it is formed by 20 different instances introduced 40 times each. This is done to perform the Reinforce with self-competing baseline described in the paper (more detailed information available in Appendix~\ref{reinforce}). The optimizer is Adam \cite{kingma2014adam} with a learning rate of $5 \cdot 10^{-4}$. The gradients are clipped to the norm by a value of 1, and a dropout with a probability of 0.1 is used in the LSTM encoder.

\begin{figure*}[t]
\centering
  \subfloat[4x4 OR-Tools Makespan=24]{%
    \includegraphics[width=.33\textwidth]{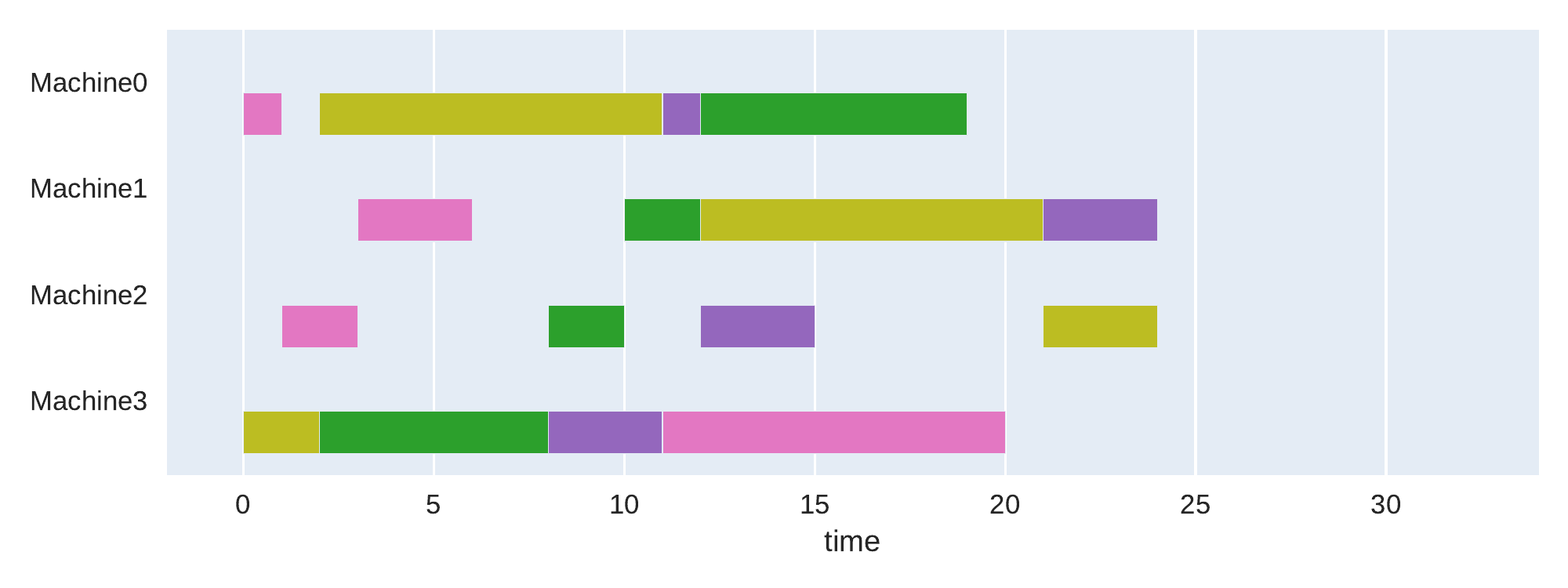}}\hfill
  \subfloat[4x4 GA Makespan=24]{%
    \includegraphics[width=.33\textwidth]{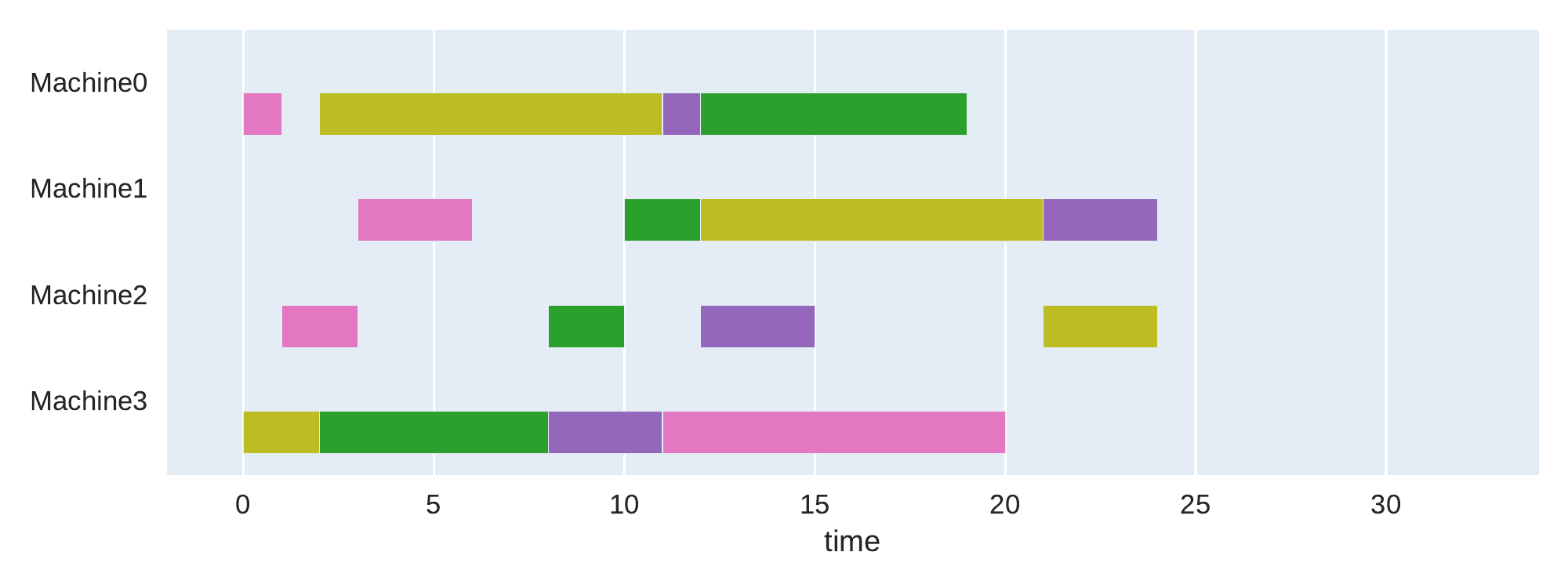}}\hfill
  \subfloat[4x4 RL Makespan=24]{%
    \includegraphics[width=.33\textwidth]{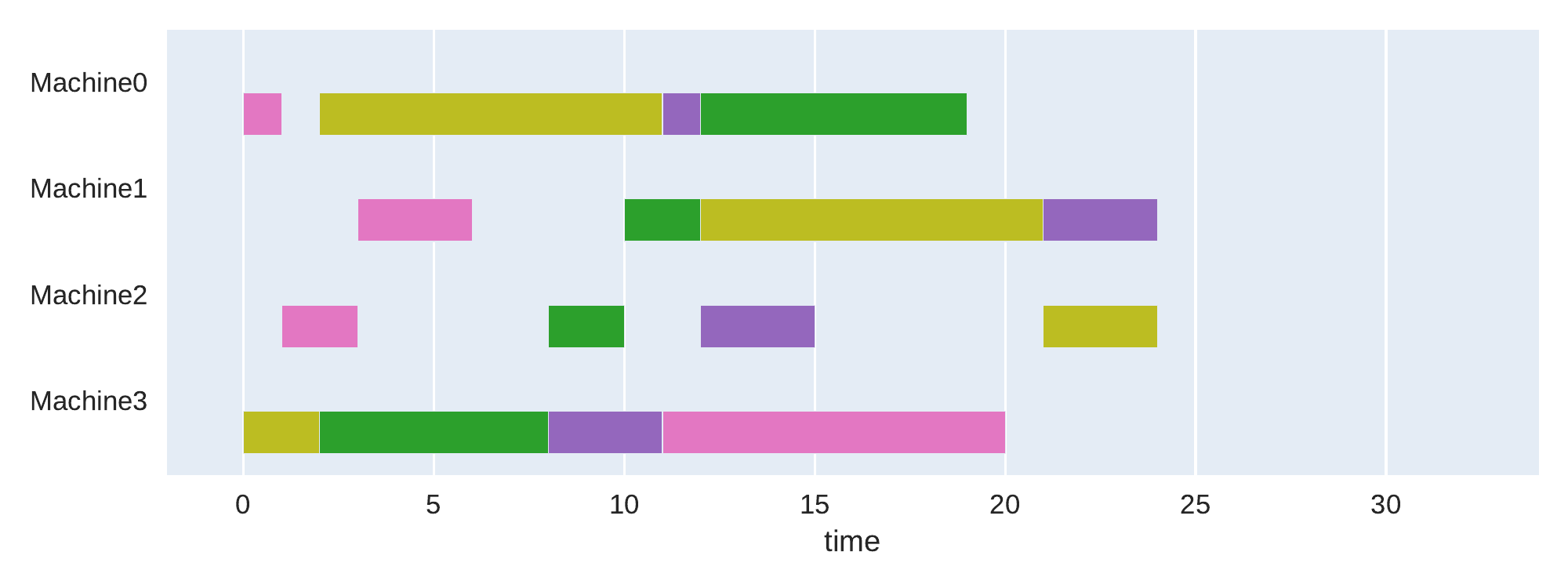}}\\
  \subfloat[10x10 OR-Tools Makespan=80]{%
    \includegraphics[width=.33\textwidth]{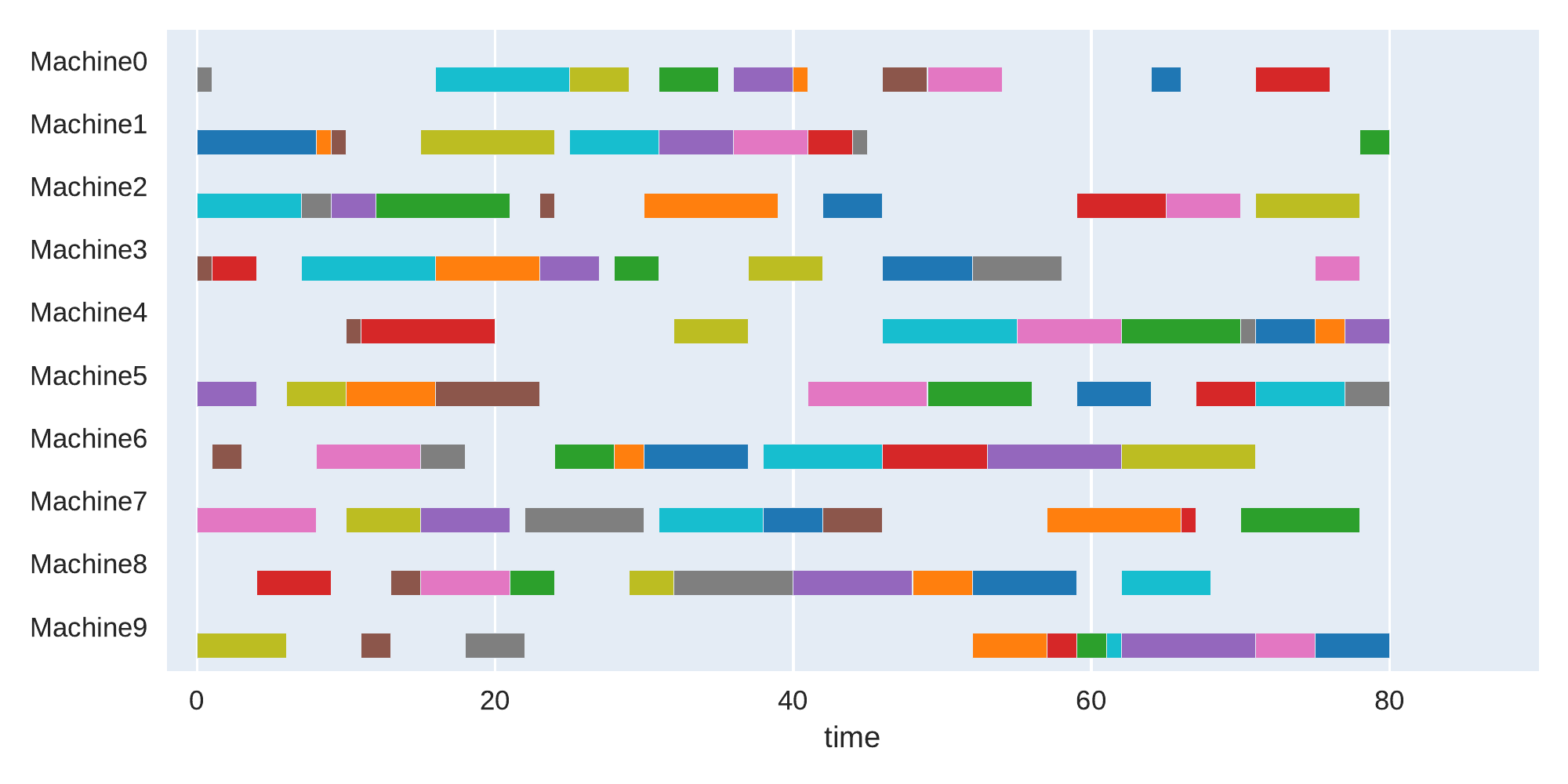}}\hfill
  \subfloat[10x10 GA Makespan=92]{%
    \includegraphics[width=.33\textwidth]{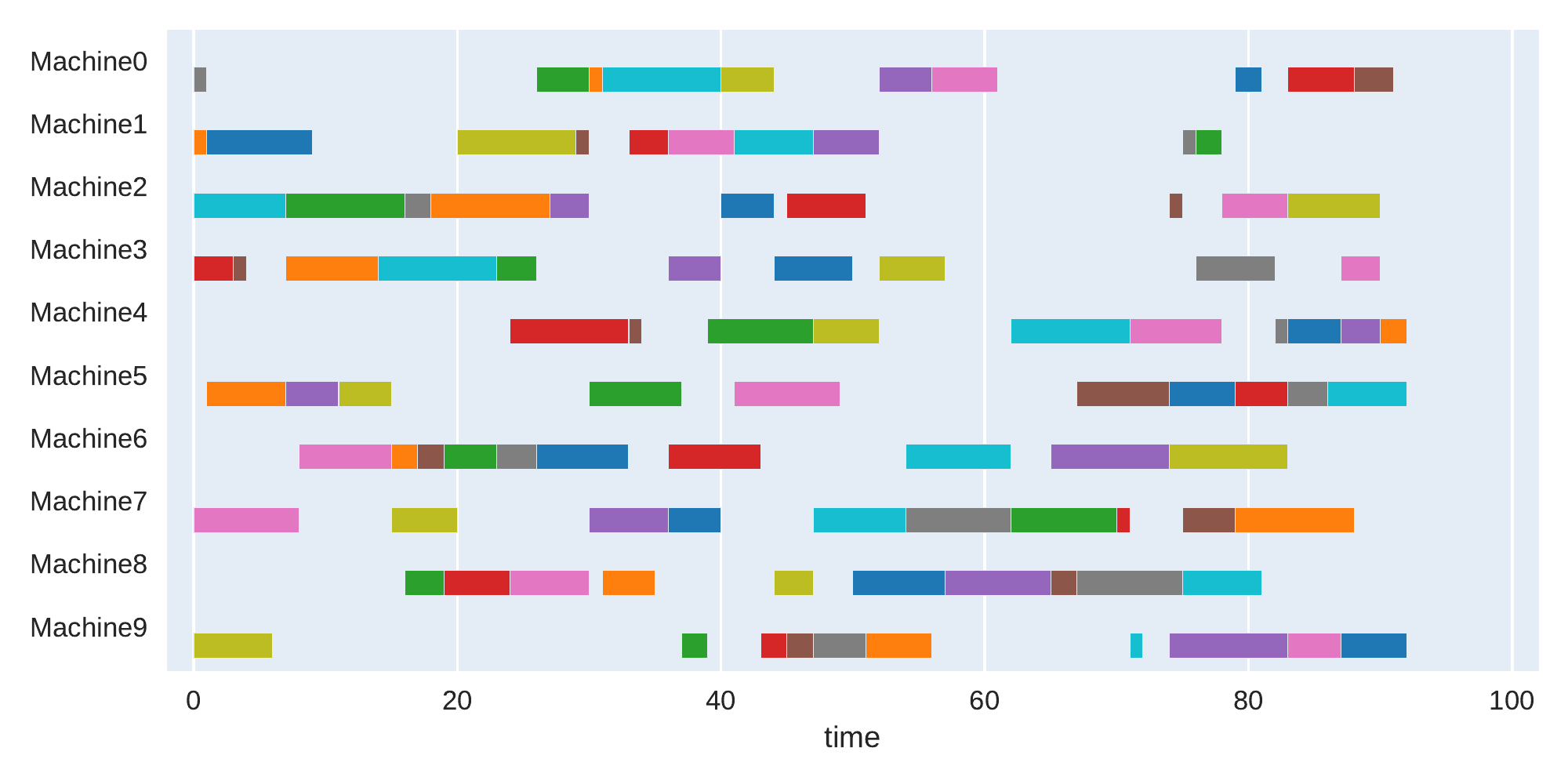}}\hfill
  \subfloat[10x10 RL Makespan=87]{%
    \includegraphics[width=.33\textwidth]{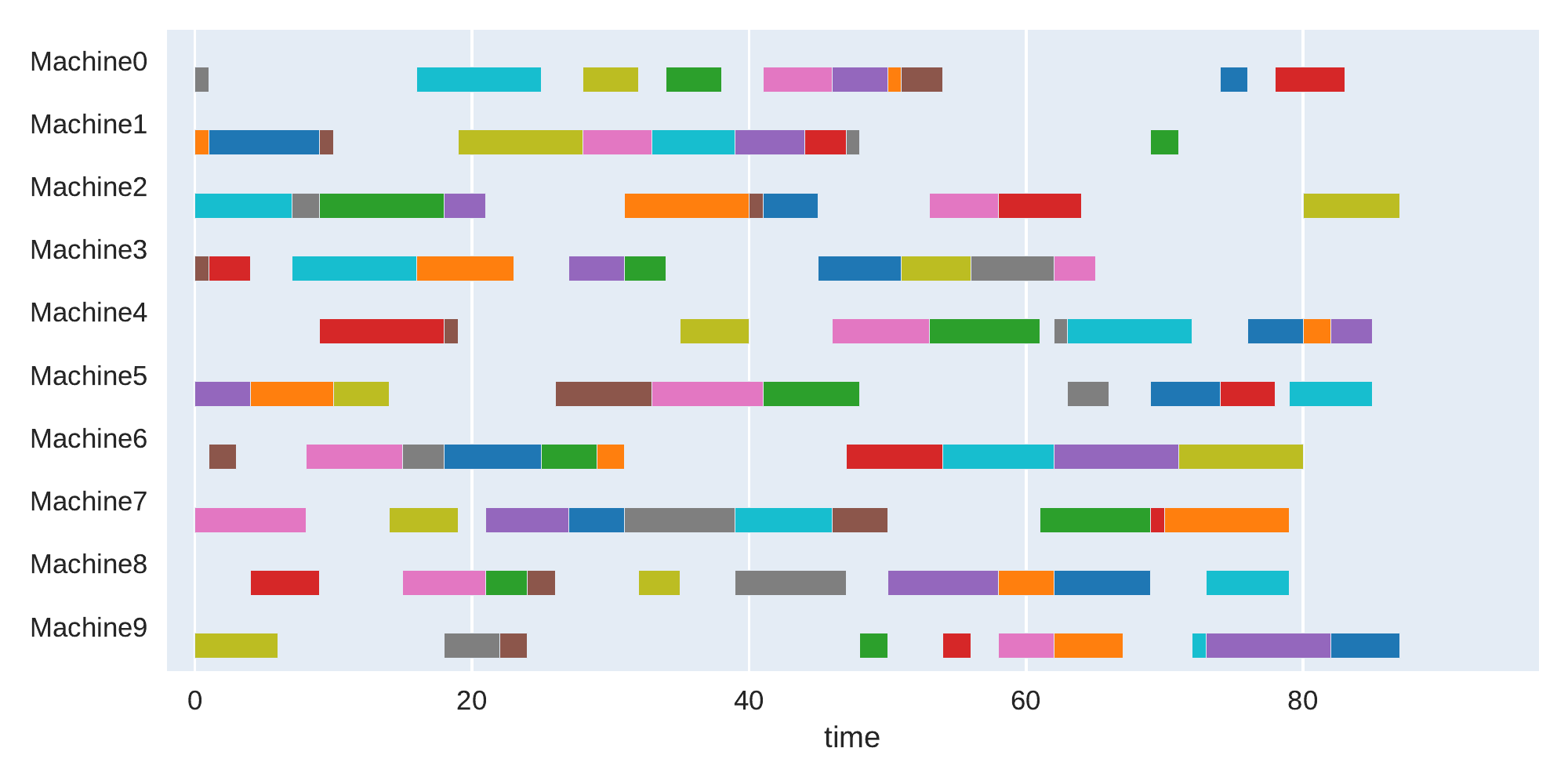}}\\
  \subfloat[15x15 OR-Tools Makespan=121]{%
    \includegraphics[width=.33\textwidth]{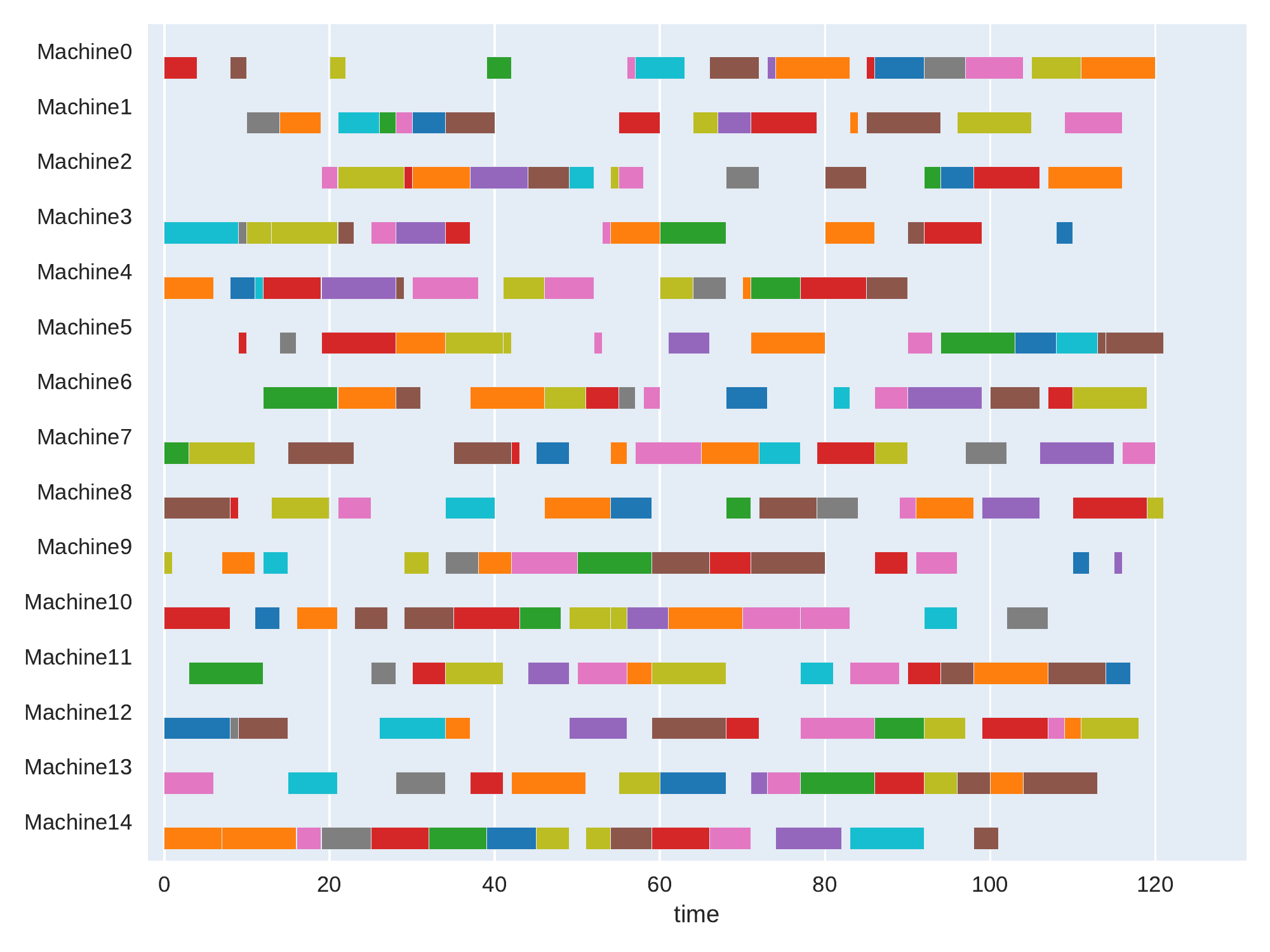}}\hfill
  \subfloat[15x15 GA Makespan=168]{%
    \includegraphics[width=.33\textwidth]{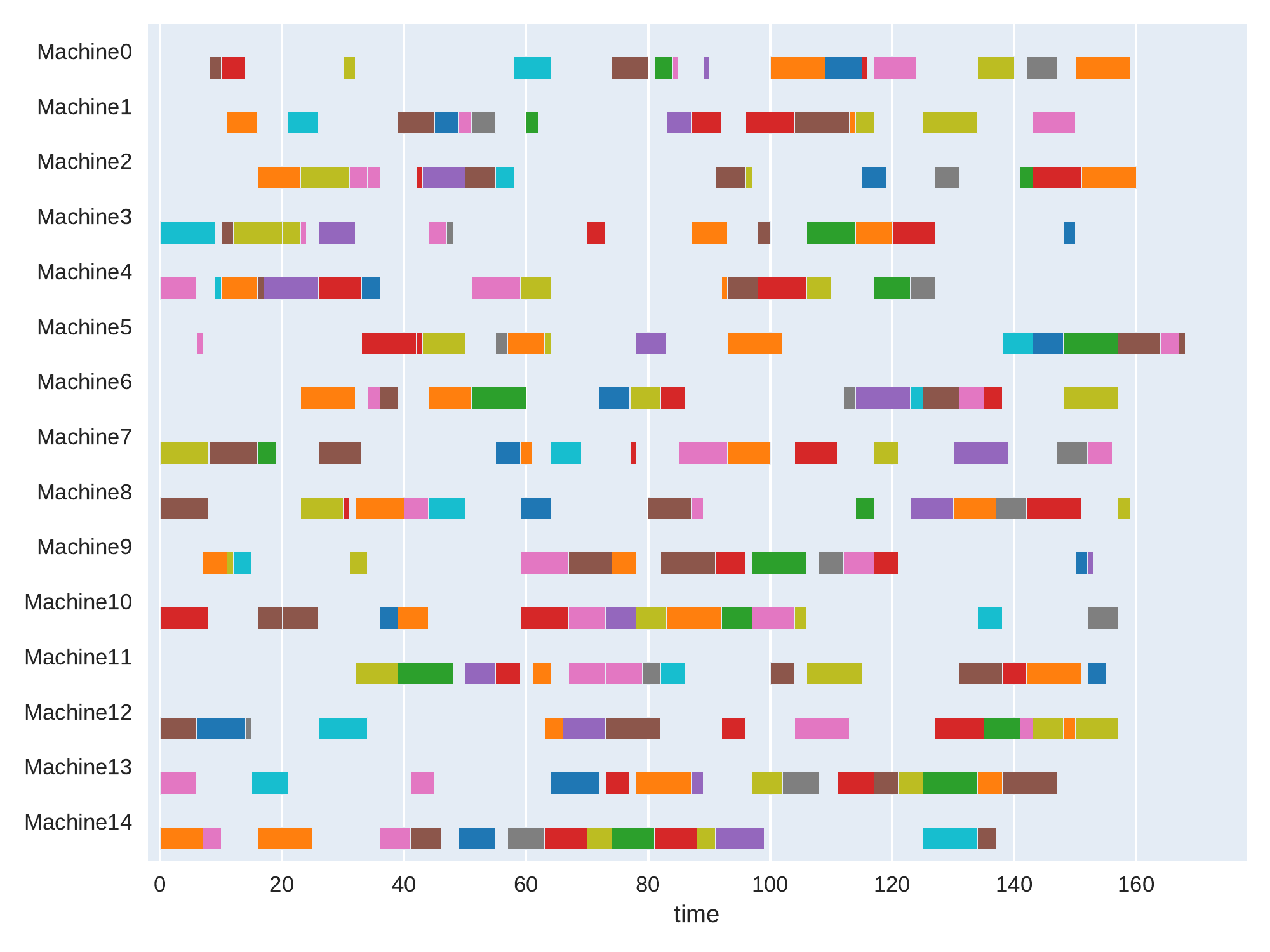}}\hfill
  \subfloat[15x15 RL Makespan=139]{%
    \includegraphics[width=.33\textwidth]{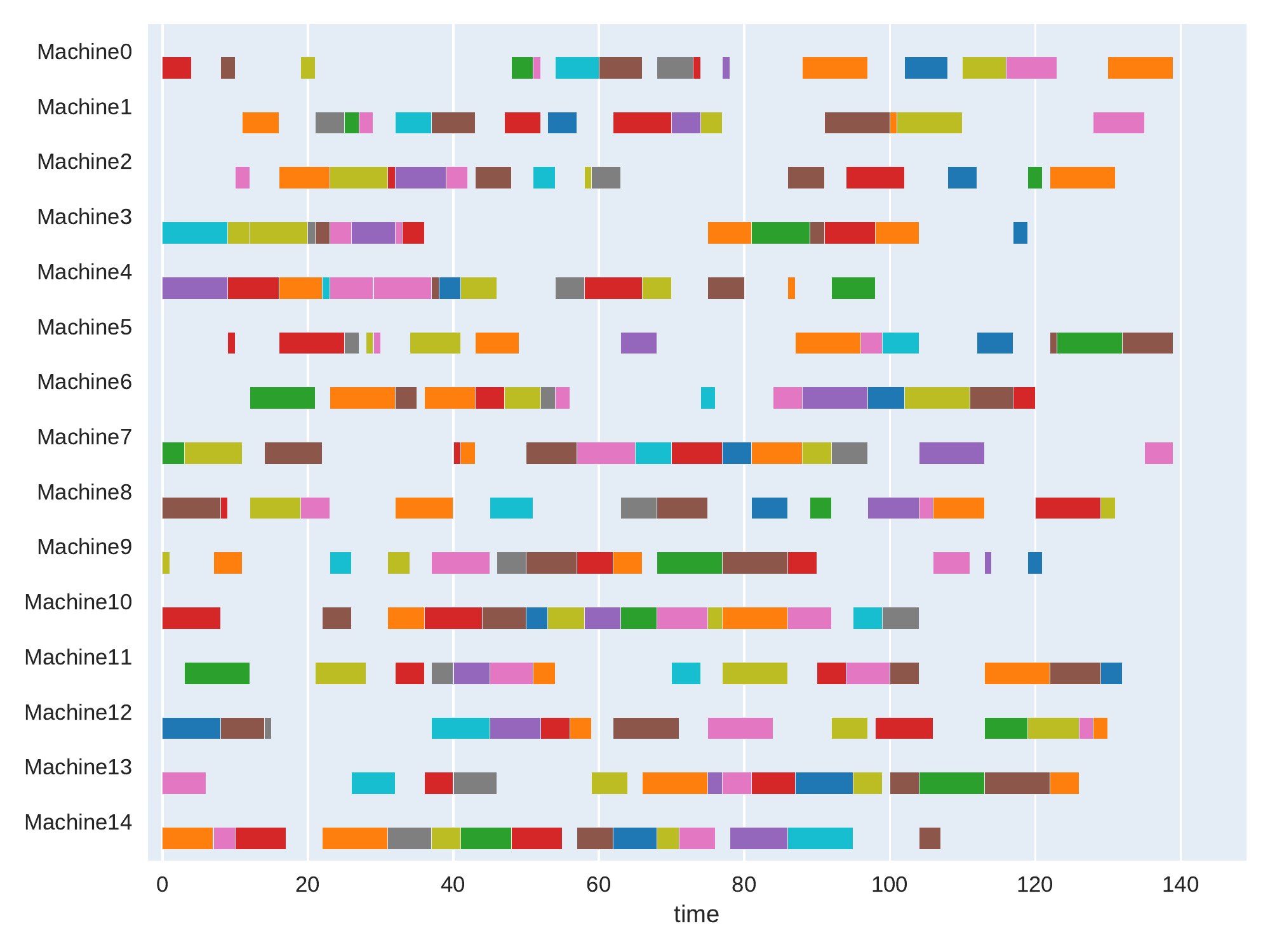}}\\

  \caption{Example solutions for the classic JSP ($\lambda=0$) problem, Gantt diagrams. On the left, the optimal solutions computed with OR-Tools; on the center, the result of the Genetic Algorithm and on the right, the results obtained by the RL model with a sampling technique.}\label{fig:1}
\end{figure*}

\subsection{Run times}

The code for the RL model proposed in the work is implemented in PyTorch\footnote{Code will be available in:  {\url{https://github.com/OptMLGroup/CCO-RL}}.} \cite{paszke2017automatic}. The model entirely run on a GPU, i.e. both the environment and the agent are implemented as tensor operations. This allows to fully parallelize the process, executing the whole batch operations at once. Even though current RL frameworks (e.g. OpenAI baselines~\cite{dhariwal2017openai}) allow to execute the environment in parallel threads using multiple CPUs, this approach permits to significantly reduce the learning time. In order to train the model a single GPU (2080Ti) was used. The times required to perform a single epoch are described below in Table~\ref{jsp-table}.

The datasets used in the experimentation are included along the code. The instances have been created following the OR-Library \cite{beasley1990or} format. For every instance, there is a heading that indicates the number of jobs $n$ and the number of machines $m$. Then, there is a one line for each job, listing the machine number and processing time for each operation. The results provided in the experimentation are obtained after performing a training of 4000 epochs on those datasets. 

\begin{table}[h]
\caption{Computation time per epoch required by the RL model in the JSP problem.}
\label{jsp-table}
\vskip 0.15in
\begin{center}
\begin{small}
\begin{sc}
\begin{tabular}{lcccc}
\toprule
  & JSP10x10 & JSP15x15 & JSP20x20 & JSP25x25 \\
\midrule
$\lambda=0$ & 2.2s & 4.7s & 7.8s & 11.5s\\
$\lambda=1$ & 2.5s & 5.4s & 9.9s & 12.7s\\
\bottomrule
\end{tabular}
\end{sc}
\end{small}
\end{center}
\vskip -0.1in
\end{table}

Finally, to visualize the results obtained by the different alternatives, a comparison of the solutions presented as Gantt diagrams is also included. This is done for the classic JSP (Figure \ref{fig:1}) and for the \textit{no idle time} variant (Figure \ref{fig:2}). Although in the figures a strategy cannot be seen at a glance, the RL model infers a competitive policy. This policy cannot be predicted, and guarantees in the results cannot be given. Yet a consistency in the results is observed.

\begin{figure*}[t]
\centering
  \subfloat[4x4 OR-Tools Objective=33]{%
    \includegraphics[width=.33\textwidth]{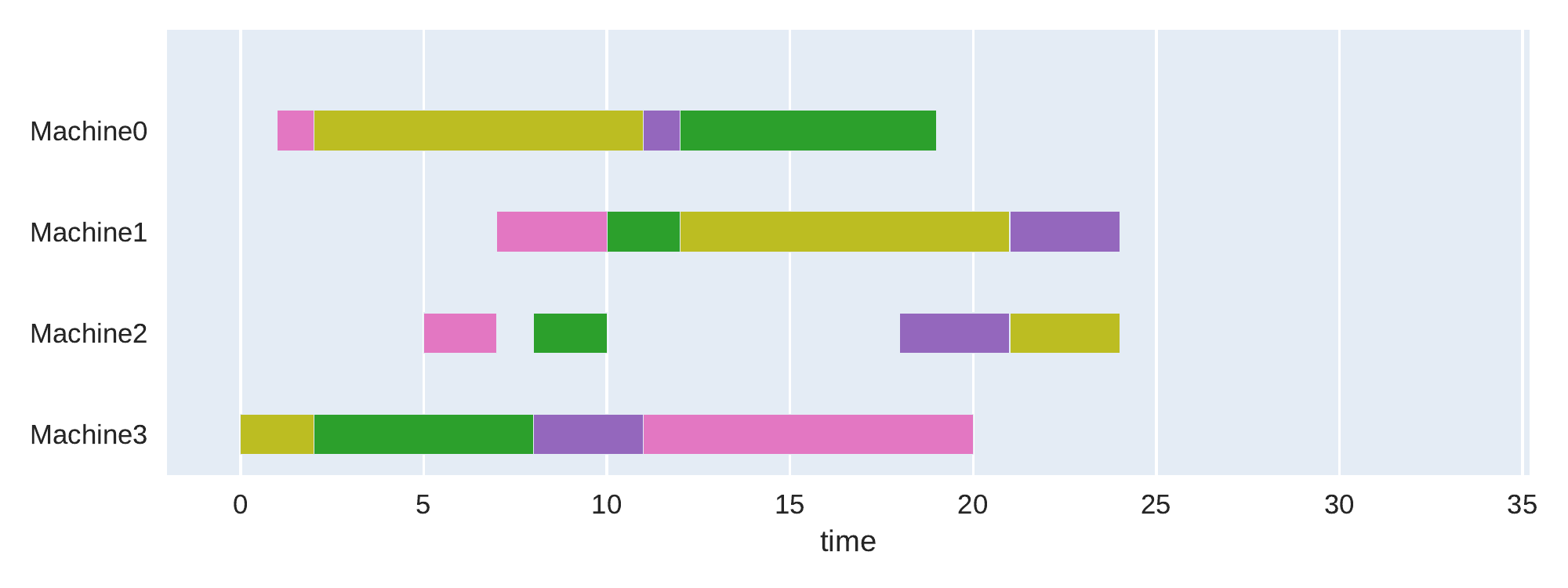}}\hfill
  \subfloat[4x4 GA Objective=33]{%
    \includegraphics[width=.33\textwidth]{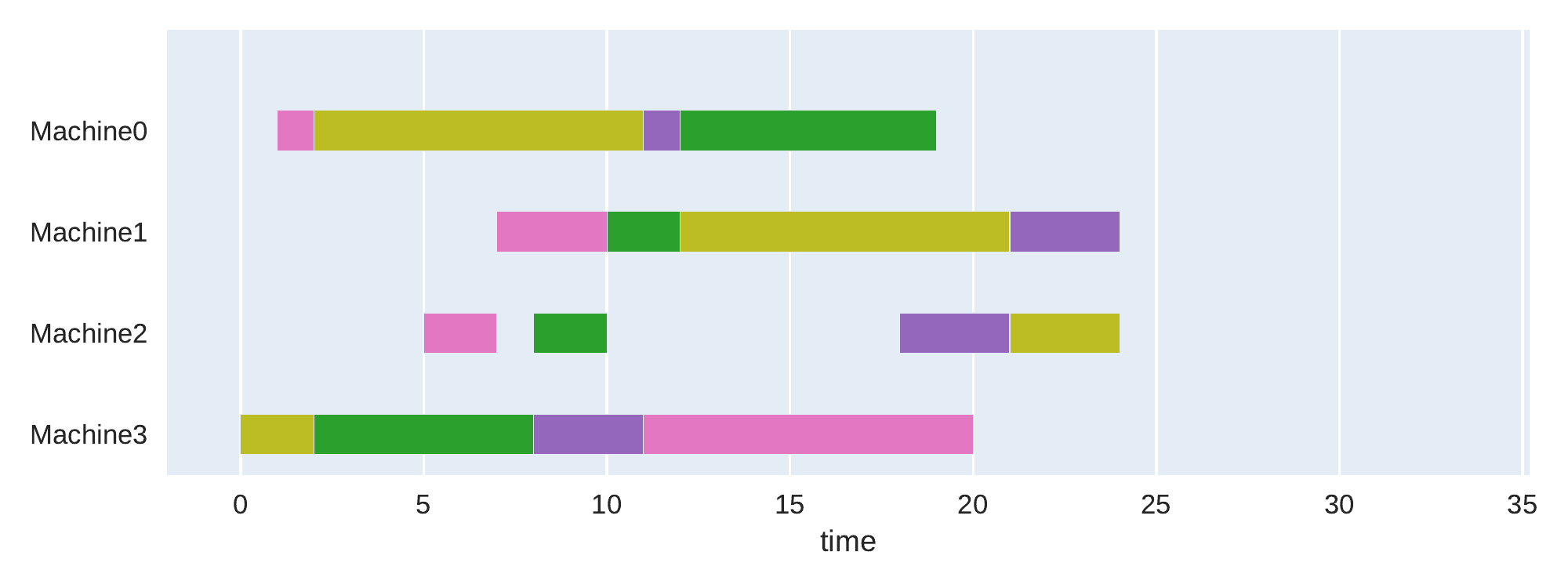}}\hfill
  \subfloat[4x4 RL Objective=45]{%
    \includegraphics[width=.33\textwidth]{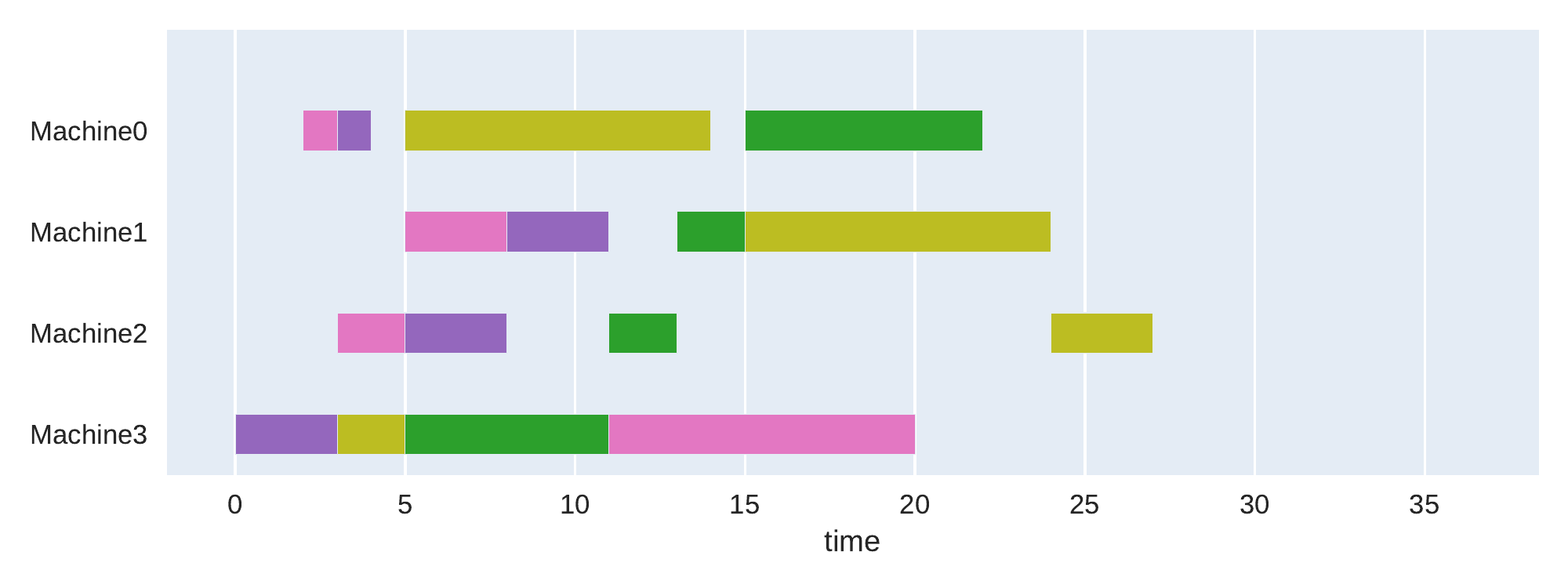}}\\
  \subfloat[10x10 OR-Tools$^{1H^{(*)}}$ Objective=236]{%
    \includegraphics[width=.33\textwidth]{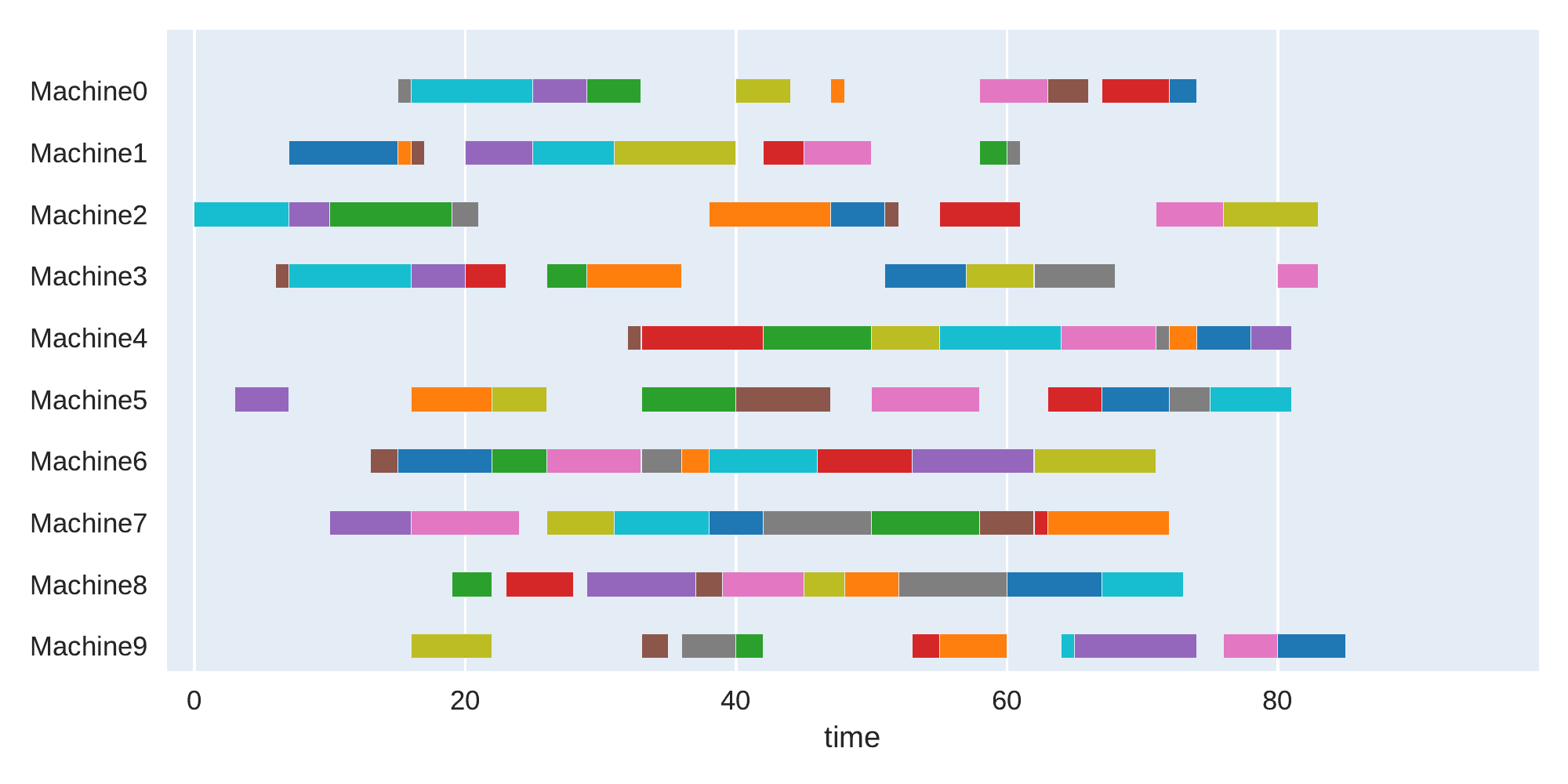}}\hfill
  \subfloat[10x10 GA Objective=318]{%
    \includegraphics[width=.33\textwidth]{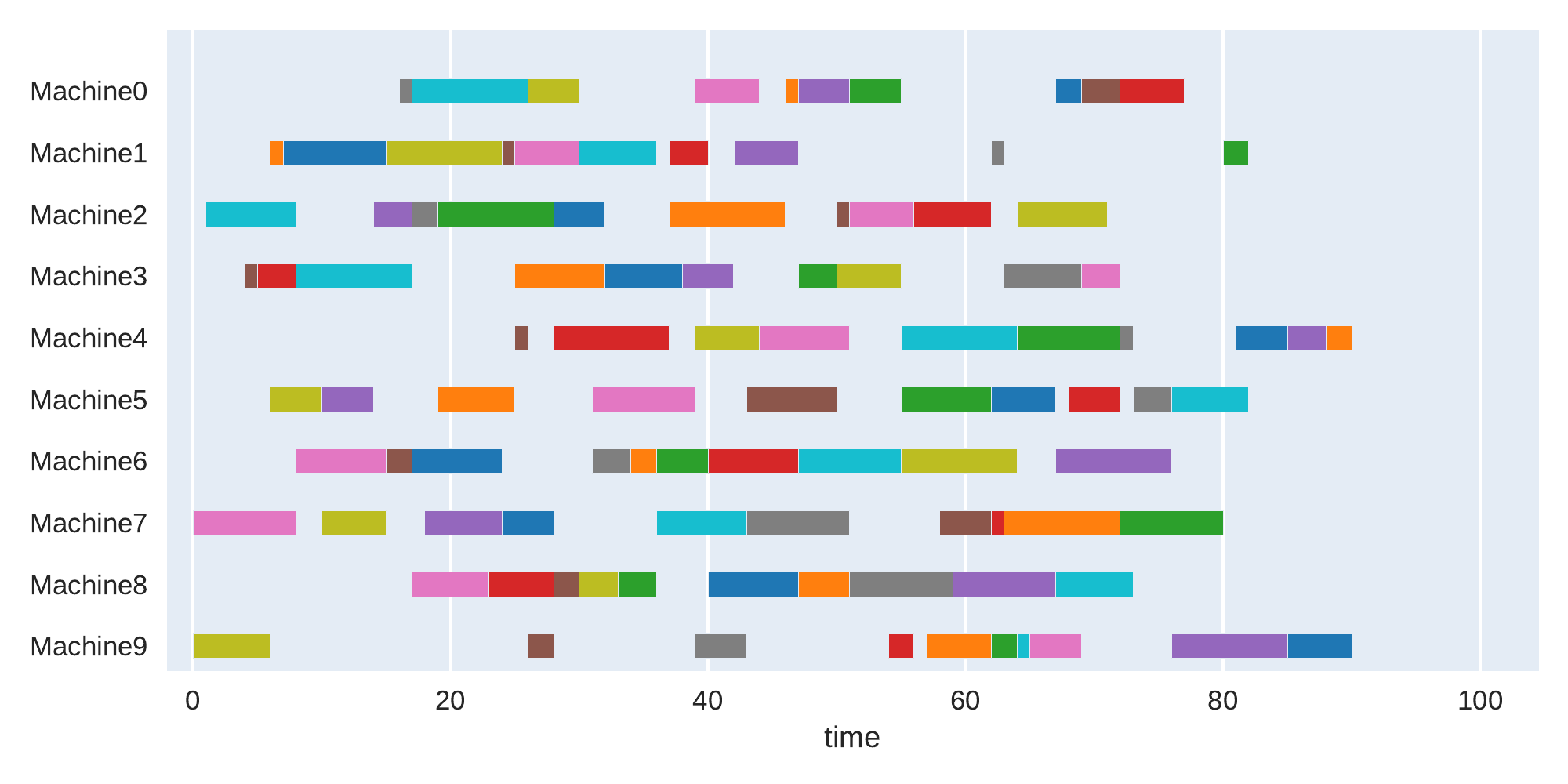}}\hfill
  \subfloat[10x10 RL Objective=392]{%
    \includegraphics[width=.33\textwidth]{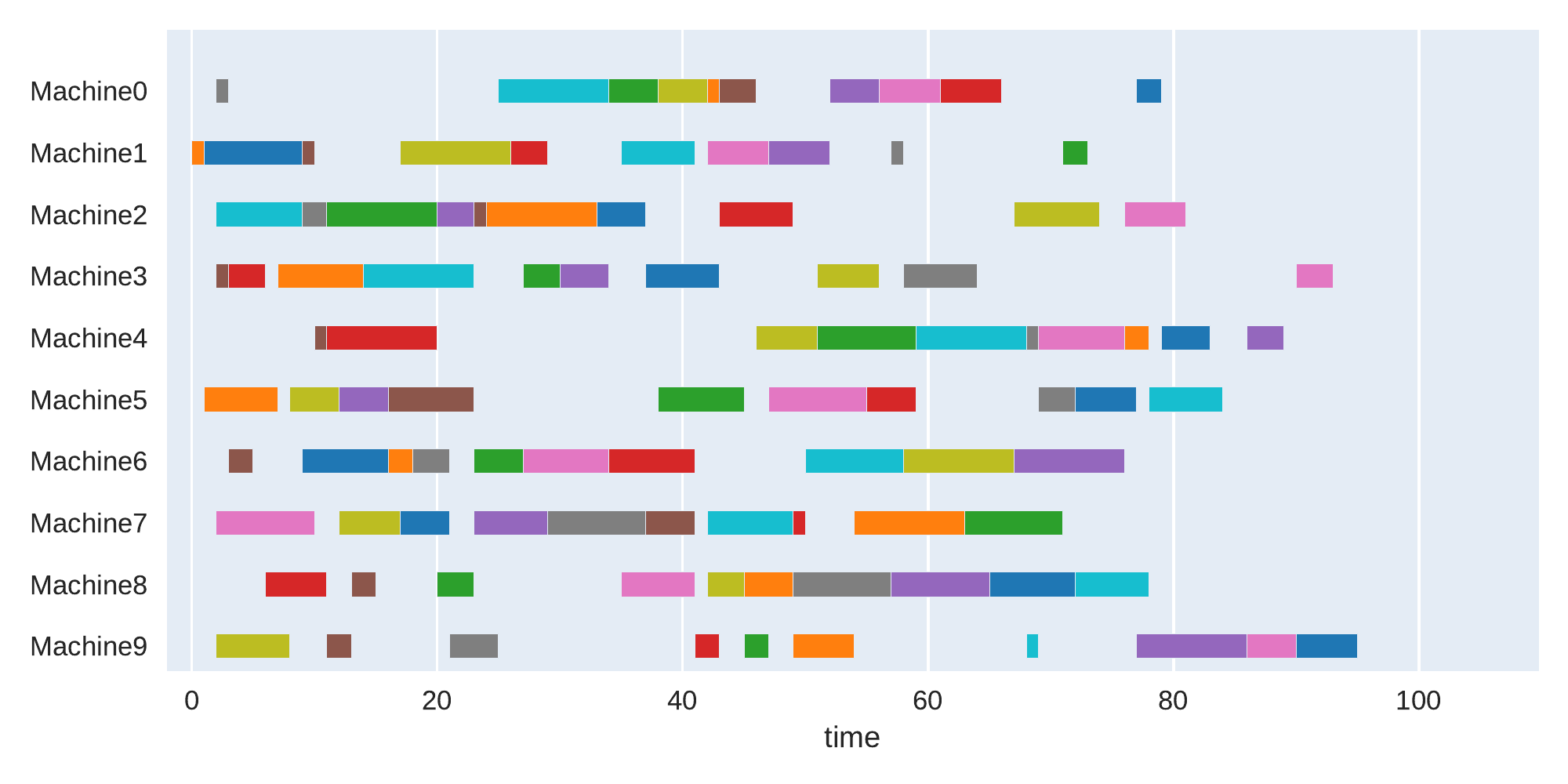}}\\
  \subfloat[15x15 OR-Tools$^{1H^{(*)}}$ Objective=588]{%
    \includegraphics[width=.33\textwidth]{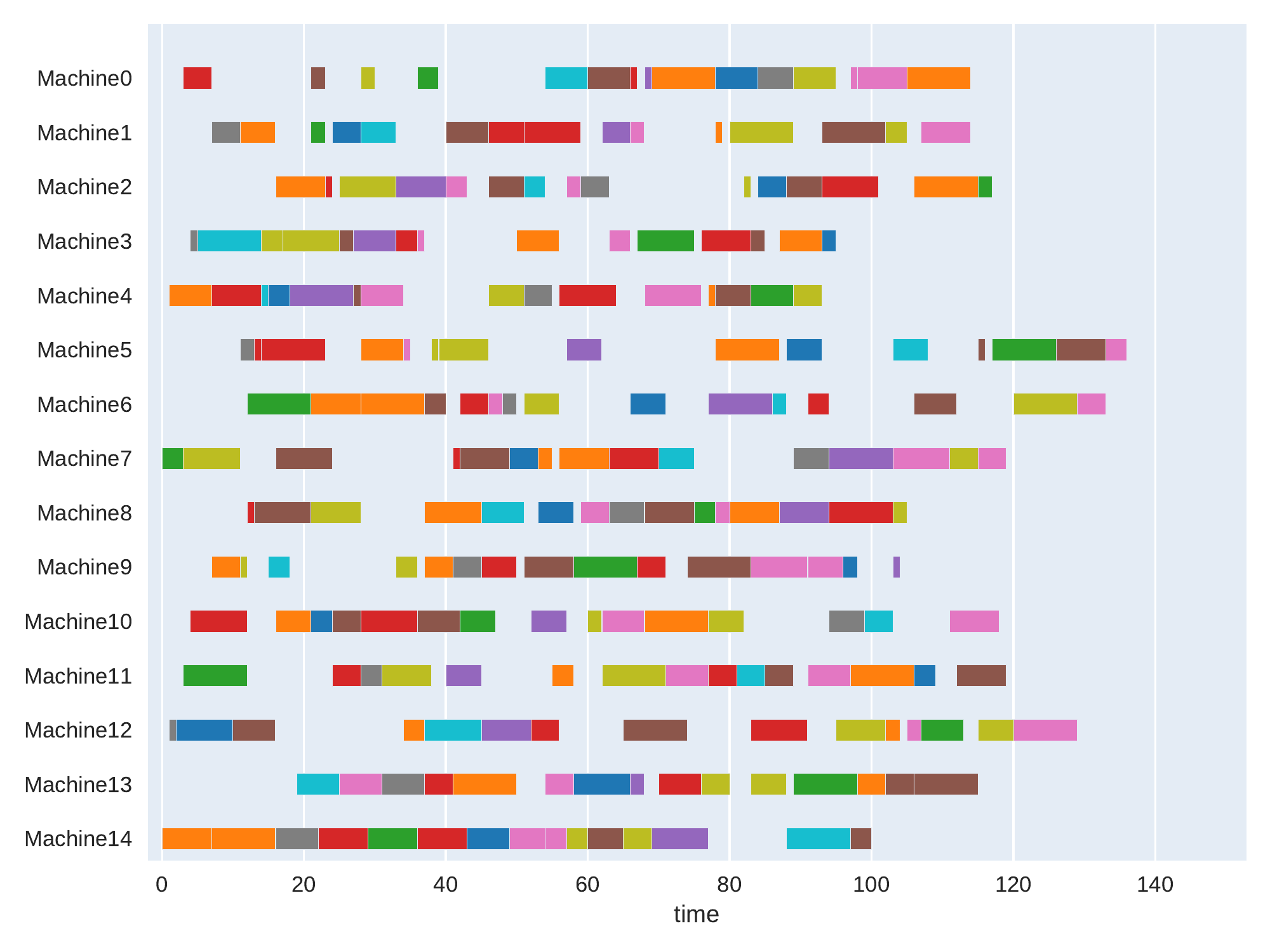}}\hfill
  \subfloat[15x15 GA Objective=1067]{%
    \includegraphics[width=.33\textwidth]{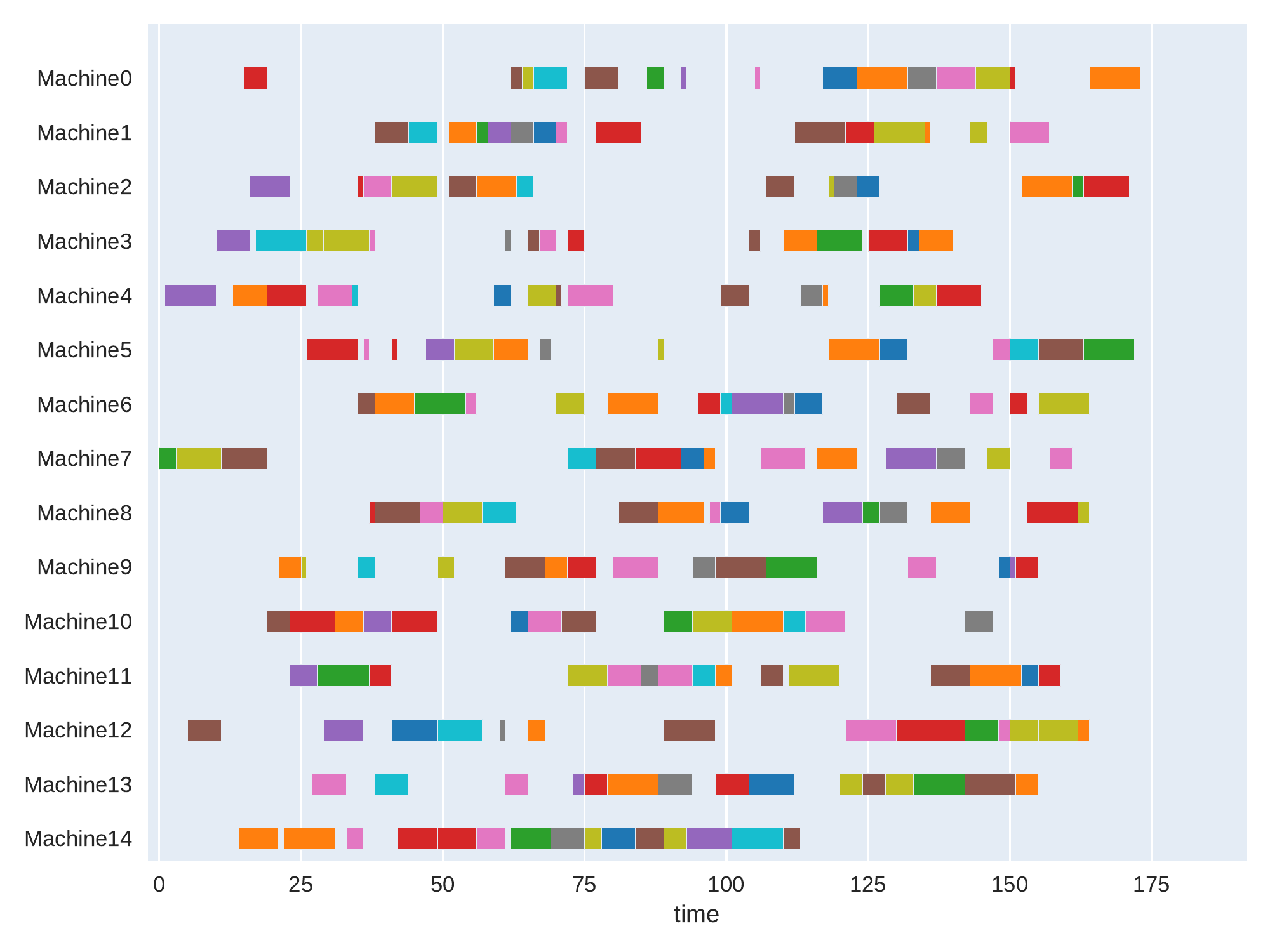}}\hfill
  \subfloat[15x15 RL Objective=868]{%
    \includegraphics[width=.33\textwidth]{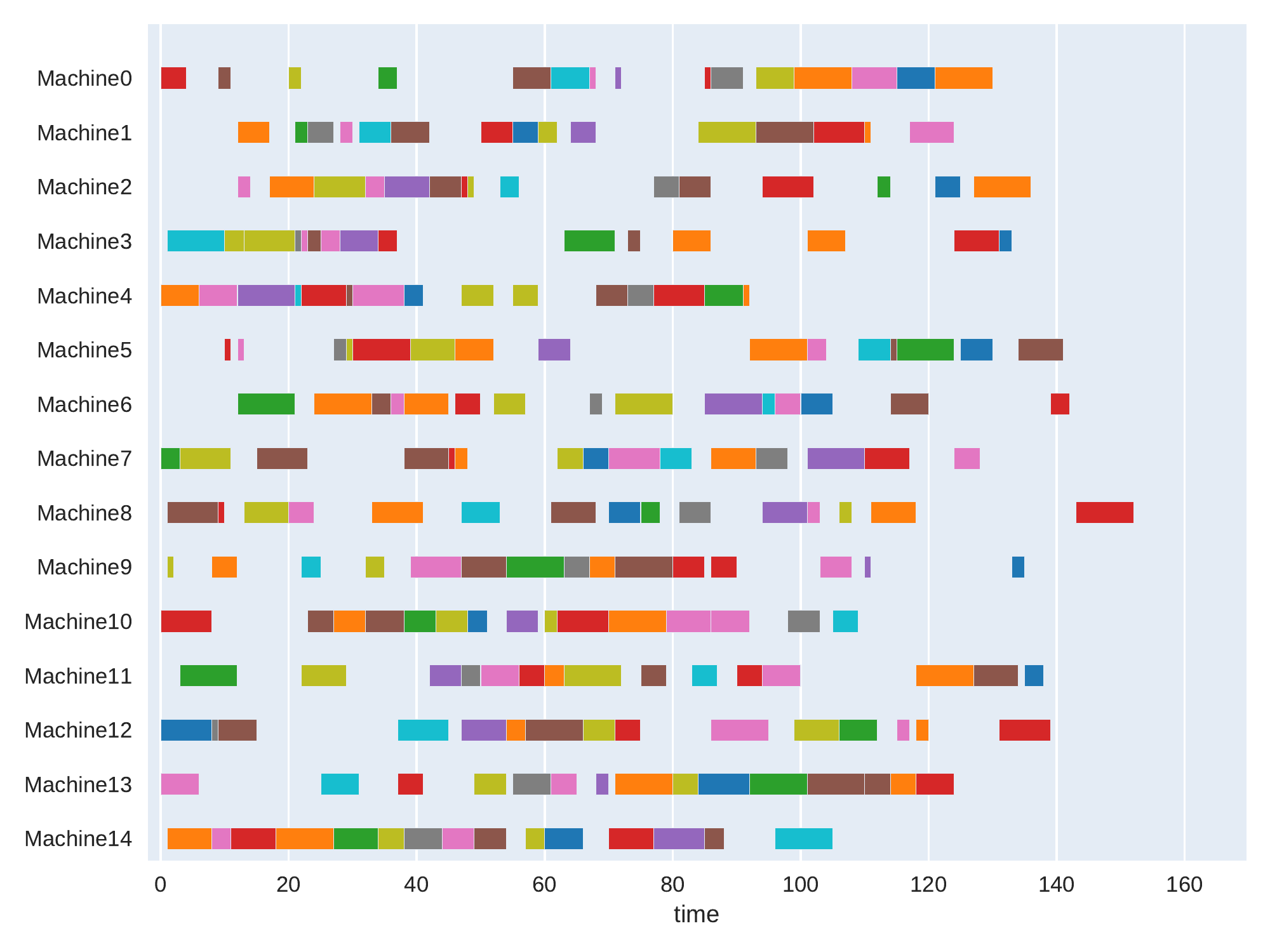}}\\

  \caption{Example solutions for the JSP \textit{with limited idle time} ($\lambda=1$) problem, Gantt diagrams. On the left, the solutions computed with OR-Tools; on the center, the result of the Genetic Algorithm and on the right, the results obtained by the RL model with a sampling technique.}\label{fig:2}
\end{figure*}

\section{Virtual Resource Allocation Problem}
\label{virtual_resource_allocation}

This appendix \ref{virtual_resource_allocation} completes the details on the Virtual Resource Allocation Problem (VRAP). The problem has been briefly introduced in Section 5.2, and in the following a complete definition and details on the neural model implementation are given.

\subsection{Problem formalization}

In the Virtual Resource Allocation Problem (VRAP), a service is required to be allocated in a pool of server hosts $H=\{H_0,H_1,...,H_{n-1}\}$. A service is defined as a unidirectional chain where the information flows from an entering Virtual Machine (VM) up to the ending machine. The service is composed by $m$ VMs selected from a service dictionary $V=\{V_0,V_1,...,V_{d-1}\}$. The order in which the information flows in the chain, $c=\{f_1,f_2,...f_m\}$ being $f \in V$, is declared in its definition. In the particular case considered in this work, server nodes are interconnected in a star configuration. 

The allocation of the services is subject to several constraints. The resources assigned to each server $H_i$ cannot exceed the resources available (number of cores $H^{cpu}_i$). In addition, the sum of ingress/egress bandwidth required by the virtual machines allocated in a server cannot exceed its bandwidth capabilities $H^{bw}_i$. A service function $f$ is defined by the number of cores $V^{cpu}_f$ it requires to run, and the bandwidth $V^{bw}_f$ of the flow it processes. Consecutive VMs that are co-located in the same server are internally connected and do not require bandwidth expenses. Finally, the problem also presents constraints related to the service itself. In the particular case considered in the work, a latency threshold has been defined, that is the sum of the computation latency $V^{lat}_f$ and networking latency associated to each link $H^{lat}_i$ cannot exceed the service agreement $L_{th}$.

The optimization problem consist of minimizing the objective function that measures the energy cost of the entire set-up. Specifically, it is calculated as the sum of the energy required to power up the servers $W^{min}$ plus the energy consumption per CPU in usage $W^{cpu}$ and networking utilization $W^{net}$. Penalties related to the service latency are introduced into the objective. Hence, the objective function is
\begin{equation}
\label{eq:1}
L = \sum_{i \in H} \biggl[W^{cpu} \cdot \sum_{f \in V} x_{fi} \cdot V^{cpu}_f + W^{min} \cdot y_{i} + W^{net} \cdot \sum_{f \in V} x_{fi} \cdot V^{bw}_f\biggl] + \lambda\biggl[\sum_{f \in V}( V^{lat}_f +  \sum_{i \in H}x_{fi} \cdot H_i^{lat}) -L_{th}\biggl]^+
\end{equation}
where  $y_{i}$ and $x_{fi}$ stand for binary decision variables indicating whether a host $i$ is activated, and the VM $f$ is located in the host respectively.

\subsection{Implementation details}

The implementation of the neural model used in the VRAP problem is similar to that 
indicated in the JSP except for small details. In this problem, instead of dealing with multiple sequences, the encoder operates with a single sequence that represents the service chain to be allocated. The iteration process has a fixed number of steps, which corresponds to the length of the service. Finally, the \textit{glimpse} mechanism operating over the encoded sequence works synchronously with the decoding mechanism. The context vector $c_t$ points at the same position over the input, the decoder is working on the output.

With regard to the parameters of the neural model, the only relevant change is related to the size of the LSTM encoder. Given the relative small length of the input sequences $m=5$, a hidden size of 16 is enough to code the information of the service chain.

\subsection{Experimentation details}

In the experimentation three different environments with 10, 20 and 50 host servers are used. Resources in those environments are initially occupied following a uniform distribution. As has been formulated in the problem, a service chain of $m=5$ elements is required to be allocated in each of them minimizing Eq.\ref{eq:1}. The VMs that conform the chain are chosen for a dictionary of 10, 20 and 50 elements respectively. We refer to these environments as VRAP10, VRAP20 and VRAP50 in the paper.

\subsection{Run times}

The run times in the VRAP are considerably shorter than the presented in the JSP example. This is due two main factors: firstly, the size of the sequences, which determine the number of iterations with the environment, is much shorter in the VRAP; and secondly, the number of parameters used in the neural model is considerably lower. As a result, the computation time required to perform a single epoch in the different scenarios are the following ones:

\begin{table}[h]
\caption{Computation time per epoch required by the RL model in the VRAP problem.}
\label{sample-table}
\vskip 0.15in
\begin{center}
\begin{small}
\begin{sc}
\begin{tabular}{lcccc}
\toprule
  & VRAP10 & VRAP20 & VRAP50 \\
\midrule
$\lambda=1$ & 0.012s & 0.017s & 0.022s \\
\bottomrule
\end{tabular}
\end{sc}
\end{small}
\end{center}
\vskip -0.1in
\end{table}

\section{Policy Gradients with self-competing baseline}
\label{reinforce}

In regard with the learning method, in the Section 3 of the paper has been argued that the learning algorithm used to implement the reward constrained policy optimization is Monte-Carlo Policy Gradients, also known as Reinforce algorithm~\cite{williams1992simple}. Here, its implementation is presented together with the self-competing baseline introduced. As argued in this work, a single neural network is employed to learn a policy $\pi_\theta$ that acts as an heuristic for solving constrained combinatorial problems. In the learning process a set of $B$ instances are sampled from the problem distribution $\mathcal{S}$. The set is computed $N$ times to estimate the objective distribution the policy presents for each instance. This procedure allows to generate a baselines estimator relying on the current stochastic policy. We call to this method self-competing baseline, as the model reinforce the best solutions the stochastic policy gets. The algorithm is described below:

\begin{algorithm}[h]
   \caption{Reinforce with self-competing baseline}
   \label{alg:reinforce}
\begin{algorithmic}

   \STATE{Initialize the actor network with random weights $\theta$}
   \FOR{episode = 1,2,...}
   \STATE{Reset the gradients: $d\theta \gets 0$}
   \STATE Sample problem instance $s^j$ from $\mathcal{S}$ for $j \in \{1,..,B\}$
   \FOR{j = 1,...,B}
        \FOR{n = 1,...,N}
    \STATE Initialize step counter: $t \gets 0$
    \REPEAT
        \STATE Sample action $y_t^{j_n}$ from the output distribution $\pi_{\theta}(\cdot|x_t^{j_n})$
        \STATE Observe the state $d_{t}^{j_n}$
        \STATE Create the input for the next step $x_{t+1}^{j_n}=\{s^j,d_t^{j_n}\}$
        \STATE $t \gets t+1$
   \UNTIL termination condition is satisfied
        \STATE Compute the objective function $L(y^{j_n}|s^j)$
        \ENDFOR
   \STATE Create the objective distribution $Q^j$ with the $N$ samples obtained for the problem instance $s^j$
   \STATE Compute the baseline: $ b(s^j)=\{q^j: Pr(Q^j \leq q^j) = \alpha\}$
   \ENDFOR
   
    \STATE Compute the gradients: $g_{\theta} = 1/(B \cdot N) \cdot \sum_{j=1}^{B}  \sum_{n=1}^{N} (L(y^{j_n})-b(s^{j})) \cdot\nabla_{\theta} \log \pi_{\theta}(y^{j_n}|s_{j}) $

    \STATE Update the weights: $\theta \gets \textrm{Adam}(\theta,g_{\theta})$

   \ENDFOR
\end{algorithmic}
\end{algorithm}

\section{Our model versus sequence-to-sequence models}

During the experimentation a sequence-to-sequence model based on a recurrent encoder-decoder architecture was also tested. Specifically, the model was applied to solve the VRAP problem. The results were considerable worst than those obtained by the iterative alternative. Presenting even for small VRAP10 instances a noticeable larger optimality gap. In addition, sequence-to-sequence models have an architecture that is not convenient for solving some combinatorial problems, i.e. problems with temporal dependencies. For example, the JSP can be addressed using a sequence-to-sequence model that outputs a categorical distribution over the jobs. In that case, the output corresponds to a sequence indicating the job to be scheduled first. Nevertheless, this approach ignores the temporary nature of the problem, not allowing the model to learn simple heuristics as the ones presented in Section \ref{heuristics}. This leads to a model that is more difficult to train in comparison to the proposed alternative.


\end{document}